\documentclass{article}
\PassOptionsToPackage{numbers, compress, sort}{natbib}
\usepackage[final]{neurips_2021}

\usepackage[utf8]{inputenc} 
\usepackage[T1]{fontenc}    
\usepackage{hyperref}
\usepackage{url}
\usepackage{subfig}
\usepackage{booktabs} 
\usepackage{algpseudocode}
\usepackage[linesnumbered, ruled, vlined]{algorithm2e}

\usepackage{chngcntr}
\usepackage{xcolor}
\usepackage{xspace}
\counterwithin{table}{section}
\usepackage{microtype}      
\usepackage{graphicx}
\usepackage{amsfonts}       
\usepackage{nicefrac} 
\usepackage{multirow}
\newcommand{\cmtt}[1]{\textnormal{\fontfamily{cmtt}\selectfont #1}}
\usepackage{bm}
\usepackage{amsmath}
\usepackage{amssymb}
\usepackage{amsthm}
\usepackage{pifont}
\usepackage{bbm}
\usepackage{bigints}
\usepackage{esint}
\usepackage{mathrsfs}
\usepackage{wrapfig}
\usepackage{enumitem}
\usepackage{adjustbox}
\usepackage[font=small,labelfont=bf,labelsep=period]{caption}
\hypersetup{colorlinks,linkcolor={blue},citecolor={blue},urlcolor={red}}
\usepackage{natbib}

\counterwithin{table}{section}
\numberwithin{equation}{section}
\numberwithin{figure}{section}
\theoremstyle{plain}
\newtheorem{thm}{\protect\theoremname}[section]
\theoremstyle{definition}

\theoremstyle{remark}

\theoremstyle{plain}

\newcommand{\bw}{\mathbf{w}}

\def\T{{ \mathrm{\scriptscriptstyle T} }}
  \providecommand{\lemmaname}{Lemma}
  
  \providecommand{\remarkname}{Remark}
 
\providecommand{\theoremname}{Theorem}

\makeatother
\newcommand{\norm}[1]{\Vert#1\Vert}
\providecommand{\definitionname}{Definition}
\providecommand{\lemmaname}{Lemma}
\providecommand{\remarkname}{Remark}
\providecommand{\theoremname}{Theorem}

\DeclareMathOperator*{\argmin}{arg\,min}

\title{Neural Bootstrapper}

\author{
  Minsuk Shin\textsuperscript{\rm 1}\thanks{Equal Contribution}\ , Hyungjoo Cho\textsuperscript{\rm 2}\footnotemark[1]\ , Hyun-seok Min\textsuperscript{\rm 3}, Sungbin Lim\textsuperscript{\rm 4}\thanks{Corresponding Author} \\
  Department of Statistics, University of South Carolina\textsuperscript{\rm 1} \\
  Department of Transdisciplinary Studies, Seoul National University\textsuperscript{\rm 2} \\
  Tomocube Inc.\textsuperscript{\rm 3} \\
  Artificial Intelligence Graduate School, UNIST\textsuperscript{\rm 4} \\
  \texttt{sungbin@unist.ac.kr}
}

\begin{document}
\maketitle

\begin{abstract}
Bootstrapping has been a primary tool for ensemble and uncertainty quantification in machine learning and statistics. 
However, due to its nature of multiple training and resampling, bootstrapping deep neural networks is computationally burdensome; hence it has difficulties in practical application to the uncertainty estimation and related tasks.
To overcome this computational bottleneck, we propose a novel approach called \emph{Neural Bootstrapper} (NeuBoots), which learns to generate bootstrapped neural networks through single model training. 
NeuBoots injects the bootstrap weights into the high-level feature layers of the backbone network and outputs the bootstrapped predictions of the target, without additional parameters and the repetitive computations from scratch. 
We apply NeuBoots to various machine learning tasks related to uncertainty quantification, including prediction calibrations in image classification and semantic segmentation, active learning, and detection of out-of-distribution samples. 
Our empirical results show that NeuBoots outperforms other bagging based methods under a much lower computational cost without losing the validity of bootstrapping.
\end{abstract}

\section{Introduction}
    \label{sec:intro}
Bootstrapping \cite{efron1979bootstrap} or bagging \cite{breiman1996bagging} procedures have been commonly used as a primary tool in quantifying uncertainty lying on statistical inference, e.g. evaluations of standard errors, confidence intervals, and hypothetical null distribution. 
Despite its success in statistics and machine learning field, the naive use of bootstrap procedures in deep neural network applications has been less practical due to its computational intensity. 
Bootstrap procedures require evaluating a number of models; however, training multiple deep neural networks are infeasible in practice in terms of computational cost.        

To utilize bootstrap for deep neural networks, we propose a novel bootstrapping procedure called \textbf{Neu}ral \textbf{Boots}trapper (NeuBoots). 
The proposed method is mainly motivated by Generative Bootstrap Sampler (GBS) \citep{shin2020GBS}, which trains a bootstrap generator by model parameterization based on Random Weight Bootstrapping (RWB, \cite{shao1996jackknife}) framework. 
For many statistical models, the idea of GBS is more theoretically valid than amortized bootstrap \cite{nalisnick2017amortized}, which trains an implicit model to approximate the bootstrap distribution over model parameters.
However, GBS is hardly scalable to modern deep neural networks 
containing millions of parameters.

Contrary to the previous method, the proposed method is effortlessly scalable and universally applicable to the various architectures.
The key idea of NeuBoots is simple; multiplying bootstrap weights to the final layer of the backbone network and instead of model parameterization.
Hence it outputs the bootstrapped predictions of the target without additional parameters and the repetitive computations from scratch.
NeuBoots outperforms the previous sampling-based methods \cite{gal2016dropout, lakshminarayanan2017simple, nalisnick2017amortized} on the various uncertainty quantification related tasks with deep convolutional networks \cite{he2016deep, huang2017densely, kim2019scalable}.
Throughout this paper, we show that NeuBoots has multiple advantages over the existing uncertainty quantification procedures in terms of memory efficiency and computational speed (see Table \ref{tab:speed-comparison}).
\begin{table*}[t]
    \centering
    \resizebox{0.9\textwidth}{!}{

    \begin{tabular}{c|c c c c c c}
        \toprule
         & Standard Bootstrap \cite{efron1979bootstrap} & MCDrop \cite{gal2016dropout} & DeepEnsemble \cite{lakshminarayanan2017simple}& NeuBoots \\
         \midrule
         Memory Efficiency & \includegraphics[height=11pt]{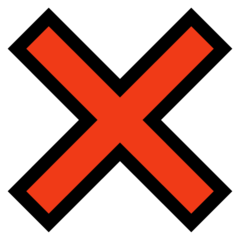} & \includegraphics[height=11pt]{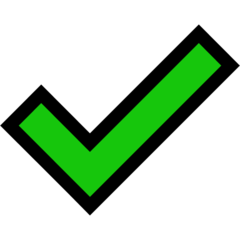} &
         \includegraphics[height=11pt]{./fig/check-no.png} & 
         \includegraphics[height=11pt]{./fig/check-yes.png} \\
         Fast Training & 
         \includegraphics[height=11pt]{./fig/check-no.png} & \includegraphics[height=11pt]{./fig/check-yes.png} &
         \includegraphics[height=11pt]{./fig/check-no.png} &
         \includegraphics[height=11pt]{./fig/check-yes.png}
         \\
         Fast Prediction & \includegraphics[height=11pt]{./fig/check-no.png} & \includegraphics[height=11pt]{./fig/check-no.png} &
         \includegraphics[height=11pt]{./fig/check-no.png} & \includegraphics[height=11pt]{./fig/check-yes.png} \\
         \bottomrule
    \end{tabular}
    }\caption{Computational comparison between bagging based uncertainty estimation methods in the view of memory efficiency and computational speed during the training and prediction step.
    }
    \label{tab:speed-comparison}
    
\end{table*}

To verify the empirical power of the proposed method, we apply NeuBoots to a wide range of experiments related to uncertainty quantification and bagging.
We apply the NeuBoots to prediction calibration, active learning, out-of-distribution (OOD) detection, semantic segmentation, and learning on imbalanced datasets.
Notably, we test the proposed method on biomedical data of high-resolution, NIH3T3 data \cite{Choi2021.05.23.445351}.
In Section \ref{sec:emp}, our results show that NeuBoots achieves at least comparable or better performance than the state-of-the-art methods in the considered applications.

\section{Preliminaries}
\label{sec:GBS}

As preliminaries, we briefly review the standard bootstrapping \cite{efron1979bootstrap} and introduce an idea of Generative Bootstrap Sampler (GBS, \cite{shin2020GBS}), which is the primary motivation of the proposed method.
Let $[m]:=\{1,\ldots,m\}$ and
denote a given training data by  $\mathcal{D}=\{(X_{i}, y_{i}):i\in[n]\}$, where each feature $X_i\in\mathcal{X}\subset\mathbb{R}^p$ and its response $y_i\in\mathbb{R}^d$. 
We denote the class of models $f:\mathbb{R}^{p}\to\mathbb{R}^{d}$ by $\mathcal{M}$.
For the standard bootstrapping, we sample $B$ sets of bootstrap data $\mathcal{D}^{(b)}=\{(X_{i}^{(b)},y_{i}^{(b)}):i\in[n]\}$ with replacement for $b\in[B]$.
 For each bootstrap data $\mathcal{D}^{(b)}$, we define a loss functional $L$ on $f\in\mathcal{M}$:
\begin{align}
    \label{eq:loss-function}
    L(f, \mathcal{D}^{(b)}):=\frac{1}{n}\sum_{i=1}^{n}\ell(f(X_{i}^{(b)}),y_{i}^{(b)})
\end{align}
where $\ell:\mathbb{R}^{d}\times\mathbb{R}^{d}\to\mathbb{R}$ is an arbitrary loss function.
Then we minimize \eqref{eq:loss-function} with respect to $f\in\mathcal{M}$ to obtain bootstrapped models: for $b\in[B]$,
\begin{align}
    \label{eq:standard-bootstrap}
    \widehat{f}^{(b)} &= \argmin_{f\in \mathcal{M}}L(f,\mathcal{D}^{(b)}).
\end{align}

\paragraph{Random Weight Bootstrapping}

It is well-known that the standard bootstrap uses only (approximately) 63\% of observations for each bootstrap evaluation \cite{lakshminarayanan2017simple}.
To resolve this problem, we use Random Weight Bootstrapping (RWB, \cite{shao1996jackknife}), which reformulates \eqref{eq:standard-bootstrap} as a sampling of bootstrapping weights for a weighted loss functional.
Let $\mathcal{W}=\{\bw\in\mathbb{R}_{+}^{n}: \sum_{i=1}^{n} w_i = n\}$ be a dilated standard $(n-1)$-simplex.
For $\mathbf{w}=(w_{1},\ldots,w_{n})\in\mathcal{W}$ and the original training data $\mathcal{D}$, we define the Weighted Bootstrapping Loss (WBL) functional on $f\in\mathcal{M}$ as follows: 
\begin{align}
    \label{eq:loss_RWB}
    L(f,\mathbf{w},\mathcal{D}):=\frac{1}{n}\sum_{i=1}^{n}w_{i}\ell(f(X_{i}),y_{i}).
\end{align}
Then for any resampled dataset $\mathcal{D}^{(b)}$, there exists a unique $\mathbf{w}\in\mathcal{W}$ such that \eqref{eq:loss-function} matches to \eqref{eq:loss_RWB}.
This reformulation provides a relaxation method to consider full data set without any omission in bootstrapping.
Precisely, 
as a continuous relaxation of the standard bootstrap, we use Dirichlet distribution \citep{newton1994approximate}; $\mathbb{P}_{\mathcal{W}} = n\times \text{Dirichlet}(1,\dots,1)$, where $\mathbb{P}_{\mathcal{W}}$ is a probability distribution on the simplex $\mathcal{W}$.
Hence RWB fully utilizes the observed data points, since sampled bootstrap weights $\mathbf{w}\sim\mathbb{P}_{\mathcal{W}}$ are strictly positive. Also, \cite{praestgaard1993exchangeably} showed that RWB achieves the same theoretical properties with these of the standard bootstrap i.e. $\mathbb{P}_{\mathcal{W}} = \mbox{Multinomial}(n;1/n,\dots,1/n)$ in \eqref{eq:loss_RWB}.

\paragraph{Bootstrap Distribution Generator}

Although RWB resolves the data discard problem, training multiple networks $\widehat{f}^{(1)},\ldots,\widehat{f}^{(B)}$ remains a computational problem, and one has to store the parameters of every network for prediction.
To reduce the computational bottlenecks, GBS \cite{shin2020GBS} proposes a procedure to train a generator function of bootstrapped estimators for parametric statistical models. 
The main idea of GBS is to parameterize the model parameter with bootstrap weight $\mathbf{w}\in\mathcal{W}$. 
When the GBS is applied to bootstrapping neural networks, it considers a \emph{bootstrap generator} $g:\mathbb{R}^{p}\times\mathcal{W}\to\mathbb{R}^{d}$ with parameter $\theta(\mathbf{w})$, where $d$ is the total number of neural net parameters in $g$, so that $g(X,\mathbf{w})=g_{\theta(\mathbf{w})}(X)$. 
Based on \eqref{eq:loss_RWB}, we define a new WBL functional:
\begin{align}
    \label{eq:loss-generator}
    \mathcal{L}(g,\mathcal{D})=\mathbb{E}_{\mathbf{w}\sim\mathbb{P}_{\mathcal{W}}}[L(g,\mathbf{w},\mathcal{D})],\quad L(g,\mathbf{w},\mathcal{D})=\frac{1}{n}\sum_{i=1}^{n}w_{i}\ell(g(X_{i},\mathbf{w}), y_{i})
\end{align}
Note that we use the Dirichlet distribution for $\mathbb{P}_{\mathcal{W}}$; hence the functional $\mathcal{L}(g,\mathcal{D})$ includes RWB procedure itself.
Analogous to \eqref{eq:standard-bootstrap}, we obtain the bootstrap generator $\widehat{g}$ by optimizing $\mathcal{L}(g,\mathcal{D})$:
\begin{align}
    \label{eq:GBS}
\widehat{g} = \argmin_{g\in\mathcal{M}}\mathcal{L}(g,\mathcal{D})
\end{align}
Then learned $\widehat{g}$ can generate bootstrap samples for given target data $X_{*}$ by plugging an arbitrary $\mathbf{w}\in\mathcal{W}$ into $\widehat{g}(X_{*},\cdot)$.
We refer to \cite[Section 2]{shin2020GBS} for detailed theoretical results on GBS. 

\paragraph{Block Bootstrapping}

The above bootstrap generator $g$ receives a bootstrap weight vector $\mathbf{w}$ of dimension $n$; hence its optimization via \eqref{eq:GBS} would be hurdled when the number of data $n$ is large. 
Hence we utilize a block bootstrapping  procedure to reduce the dimension of bootstrap weight vector. 
We allocate the index set $[n]$ to $S$ number of blocks. 
Let $u:[n] \to [S]$ denotes the assignment function.
Then we impose the same value of weight on all elements in a block such as,
$w_{i} = \alpha_{s}$ for $u(i)=s\in [S]$,
where $\bm\alpha=(\alpha_{1},\dots,\alpha_{S}) \sim S\times\text{Dirichlet}(1,\ldots,1)$.
Instead of $\mathbf{w}$,
we plug $\boldsymbol{\alpha}$ in $g(X,\cdot)=g_{\theta(\cdot)}(X)$ to generate bootstrap samples and compute the weighted loss function in \eqref{eq:loss-generator}:
\begin{align}
    \label{eq:block-bootstrap-loss}
    \mathcal{L}(g,\mathcal{D})=\mathbb{E}_{\boldsymbol{\alpha}\sim S\times\text{Dirichlet}(1,\ldots,1)}\left[\frac{1}{n}\sum_{i=1}^{n}\alpha_{u(i)}\ell(g(X_{i},\boldsymbol{\alpha}),y_{i})\right]
\end{align}
The above procedure asymptotically converges to the same target distribution where the conventional non-block bootstrap converges.
See appendix \ref{appendix-sec:block-boots} for more detailed procedure and proofs.

\section{Neural Bootstrapper}
\label{sec:NBT}

Now we propose \textbf{Neu}ral \textbf{Boots}trapper (NeuBoots), which reduces computational complexity and memory requirement of the networks in the learning of bootstrapped distribution to being suitable for deep neural networks.

\paragraph{How to implement the bootstrap generator $g$ for deep neural networks?}

\begin{figure}[!t]
\begin{center}
    \includegraphics[width=1\textwidth]{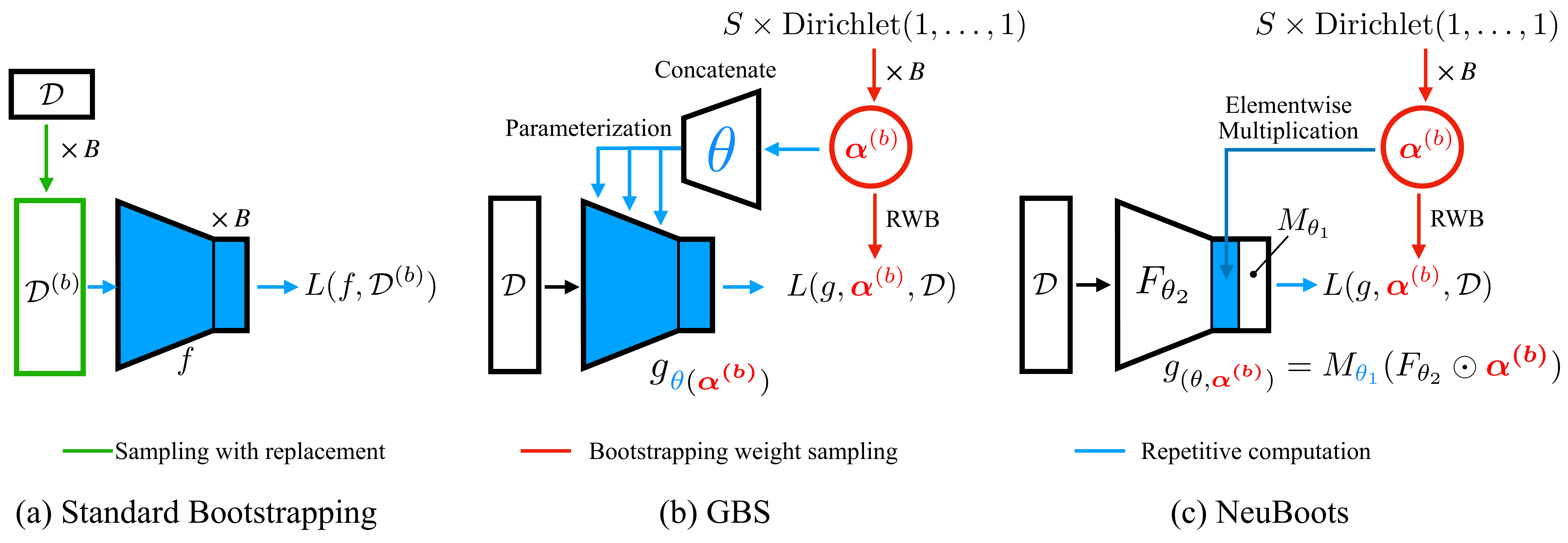}
    \caption{A comparison between the bootstrapping procedure of (a) standard bootstrapping \cite{efron1979bootstrap}, (b) GBS \cite{shin2020GBS}, and (c) NeuBoots. This figure is best viewed in color. }\label{fig:NeuBoots}
\end{center}
\end{figure}

One may consider directly applying GBS to existing deep neural networks by modeling a neural net $\theta(\cdot)$ that outputs the neural net parameters of $g$. However, this approach is computationally challenging due to the high-dimensionality of the output dimension of $\theta(\cdot)$ 
Indeed, \cite{shin2020GBS} proposes an architecture which concatenates bootstrap weight vector to every layer of a given neural network (Figure \ref{fig:NeuBoots}(b)) and trains it with \eqref{eq:block-bootstrap-loss}.
However, the bagging performance of GBS gradually degrades as we applied it to the deeper neural networks.
This may be because the information of bootstrap weights in the earlier layers less propagate since the target model reduces the parameters of the weights during the training.


\subsection{Adaptive Block Bootstrapping}
We found that the bootstrap weight in the final layer mainly affects the bootstrap performance of GBS.
This fact motivates us to utilize the following adaptive block bootstrapping, which is the key idea of NeuBoots.
Take a neural network $f_{\theta}\in\mathcal{M}$ with parameter $\theta$.
Let $M_{\theta_{1}}$ and $F_{\theta_{2}}$ be the single-layer neural network in the final layer and the feature extractor of $f$, respectively, with parameter $\theta=(\theta_{1}, \theta_{2})$, so we can decompose $f_{\theta}$ into $M_{\theta_{1}}\circ F_{\theta_{2}}$.
Set $S:=\dim(F_{\theta_{2}}(X))$ for the number of blocks for block bootstrapping.
Then, we redefine bootstrap generator as follows:
\begin{align}
    \label{eq:neuboots}
    g_{\theta}(X, \boldsymbol{\alpha}):=g_{(\theta,\boldsymbol{\alpha})}(X)=M_{\theta_{1}} (F_{\theta_{2}}(X)\odot \bm\alpha)
\end{align}
where $\odot$ denotes an elementwise multiplication. 
Bootstrap generator \eqref{eq:neuboots} can also be trained with \eqref{eq:block-bootstrap-loss}; hence optimized $\widehat{g}_{\theta}(X,\cdot)$ can generate the bootstrapped prediction as we plug $\boldsymbol{\alpha}$. 
This modification brings a computational benefit, since we can generate bootstrap samples quickly and memory-efficiently by reusing a priori computed tensor $F_{\theta_{2}}(X)$ without repetitive computation from scratch.
See Figure \ref{fig:NeuBoots} for the comparison between the previous methods and NeuBoots. 
In our empirical experience, the bootstrap evaluations over different groupings were consistent for all examined examples in this article.


\paragraph{Training and Prediction}

At every epoch, we update the $\mathbf{w}_{\boldsymbol{\alpha}}=\{\alpha_{u(1)},\ldots,\alpha_{u(n)}\}$ randomly, and the expectation in \eqref{eq:block-bootstrap-loss} can be approximated by the average over the sampled weights.
Considering the stochastic gradient descent (SGD) algorithms to update the parameter $\theta$ via mini-batch sequence $\{\mathcal{D}_{k}:\mathcal{D}_{k}\subset\mathcal{D}\}_{k=1}^{K}$, we plug mini-batch size of bootstrap weight vector $\{\alpha_{u(i)}: X_{i}\in\mathcal{D}_{k}\}$ in \eqref{eq:block-bootstrap-loss} without changing $\bm\alpha$. 
Each element of $w_{\bm\alpha}$ is not used repeatedly during the epoch, so the sampling and replacement procedures in Algorithm \ref{alg:alg-training} are conducted once at the beginning of epoch.
After we obtain the optimized network $\widehat{g}_{\theta}$, for the prediction, we use the generator $\widehat{g}_{*}(\cdot) = \widehat{g}_{\theta}(X_{*},\cdot)$ for a given data point $X_{*}$.
Then we can generate bootstrapped predictions by plugging  $\bm\alpha^{(1)},\ldots,\bm\alpha^{(B)}$ in the generator $\widehat{g}_{*}(\cdot)$, as described in Algorithm \ref{alg:alg-predict}.

\SetKwInOut{Input}{Input}
\SetKwInOut{Output}{Output}

\begin{algorithm}[!h]
\Input{Dataset $\mathcal{D}$; epochs $T$; dimension of feature $S$; index function $u$; learning rate $\rho$.}

Initialize neural network parameter $\phi^{(0)}$ and set $n:=|\mathcal{D}|$.

\For{$t\in\{0,\ldots,T-1\}$}{ 
    Sample ${\bm\alpha}^{(t)}=
    \{\alpha_{1}^{(t)},\ldots,\alpha_{S}^{(t)}\}\overset{\text{i.i.d.}}{\sim} S\times \text{Dirichlet}(1,\dots,1)$
    
    Replace $\mathbf{w}_{\bm\alpha}^{(t)} = \{\alpha_{u(1)}^{(t)},\ldots, \alpha_{u(n)}^{(t)}\}$

    Update $\theta^{(t+1)} \leftarrow \theta^{(t)}-\rho\nabla_{\theta}L(g_{(\theta,\boldsymbol{\alpha})},\mathbf{w}_{\boldsymbol{\alpha}}^{(t)},\mathcal{D})\big|_{\theta=\theta^{(t)}}$.
    }
\caption{Training step in NeuBoots.}
\label{alg:alg-training}
\end{algorithm}

\begin{algorithm}[!h]
\Input{ Data point $X_{*}\in\mathbb{R}^p$; number of bootstrap sampling $B$. }

Compute the feed-forward network $\widehat{g}_{*}(\cdot)=\widehat{g}_{\theta}(X_{*},\cdot)$ a priori.

\For{$b\in\{1,\dots,B\}$}{

Generate $\bm\alpha^{(b)}\overset{\text{i.i.d.}}{\sim} S\times \text{Dirichlet}(1,\dots,1)$ and evaluate $\widehat{y}^{(b)   }_{*}=\widehat{g}_{*}({\boldsymbol{\alpha}}^{(b)})$.
}
\caption{Prediction step in NeuBoots.}
\label{alg:alg-predict}
\end{algorithm}






\begin{figure}[h]
    \centering
    \subfloat[ Frequentist Coverage]{\includegraphics[width=0.34\textwidth]{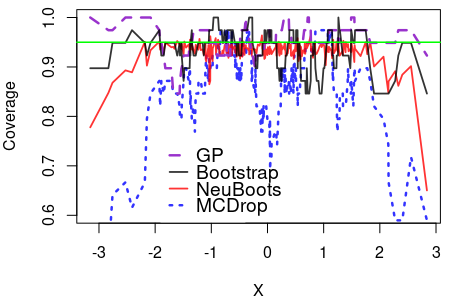} }
    \subfloat[ NeuBoots]{\includegraphics[width=0.33\textwidth]{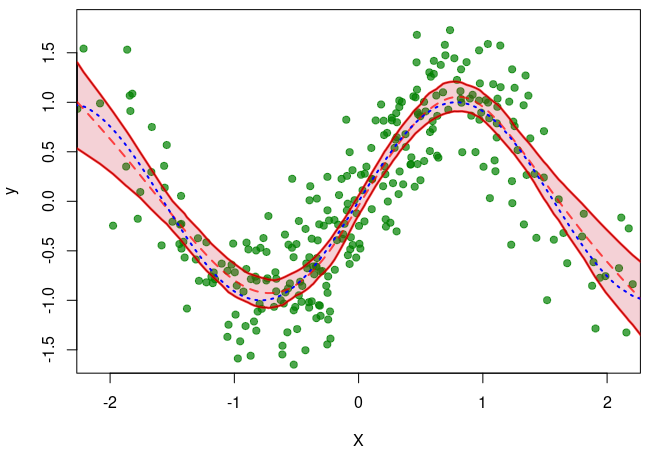} }
    \subfloat[ Standard Bootstrap]{\includegraphics[width=0.33\textwidth]{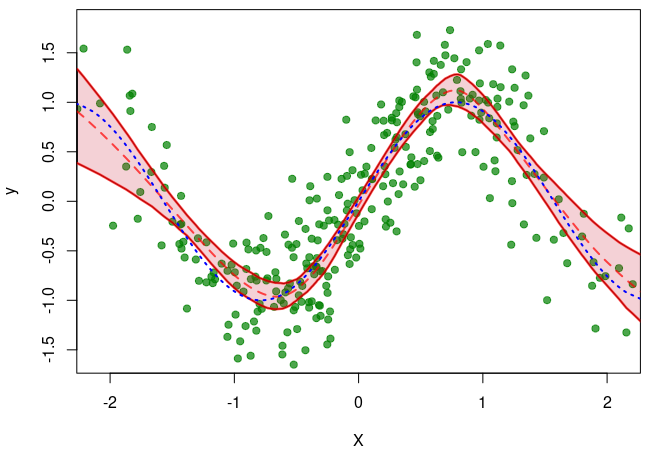} }
    \caption{(a) Frequentist coverage rate of the 95\% confidence bands; (b) Curve fitting with different nonlinear funtions. 95\% confidence band of the regression mean from NeuBoots; (c) the standard bootstrap. Each red dashed line indicates the mean, and the blue dotted lines show the true regression function.}
    \label{fig:NeuBoots-GP-MCDropout}
\end{figure}

\subsection{Discussion}
    \label{sec:regression}

\paragraph{NeuBoots vs Standard Bootstrap}

To examine the approximation power of NeuBoots, we have measured the frequentist's coverage rate of the confidence bands (Figure \ref{fig:NeuBoots-GP-MCDropout}.(a)).
We estimate 95\% confidence band for nonparametric regression function by using the NeuBoots, and compare it with credible bands (or confidence bands) evaluated by the standard bootstrap, Gaussian Process (GP) regression, and MCDrop \citep{gal2016dropout}. 
We adopt Algorithm \ref{alg:alg-training} to train the NeuBoots generator with 3 hidden-layers with 500 hidden-nodes for each layer. 
For the standard bootstrap, we train 1000 neural networks. 
The result shows the confidence band via NeuBoots stably covers the true regression function on each predictor value with almost 95\% of frequency, which is compatible with the standard bootstrapping.
In contrast, the coverage of the MCDrop is unstable and sometimes below 70\%.
This result indicates that the NeuBoots performs comparably with the standard bootstrapping in uncertainty quantification tasks.

\paragraph{NeuBoots vs Amortized Bootstrapping}

We applied NeuBoots to classification and regression experiments presented by the amortized bootstrap \cite{nalisnick2017amortized}.
Indeed, every experiment demonstrates that NeuBoots outperforms the amortized bootstrap in bagging performance for various tasks: the rotated MNIST classification (Table \ref{tab:rotatedMNIST}), classification with different data points $N$ (Figure \ref{fig:amortized-bootstrap-classification}), and regression on two datasets (Figure \ref{fig:amortized-bootstrap-regression}). 
We remark the expected calibration error (ECE, \cite{naeini2015obtaining}) score on the rotated MNIST is improved via NeuBoots from 15.00 to 2.98 by increasing the number of bootstrap sampling $B$.

\begin{table*}[h]
\footnotesize
\center
\begin{tabular}{c c c c }
\toprule
\multirow{2}{*}{Methods} & \multicolumn{3}{c}{Test Error} \\ 
& $B=1$ & $B=5$ & $B=25$
\tabularnewline
\midrule
Traditional Bootstrap & 22.57 & 19.68 & 18.57 \\
Amortized Bootstrap  & \textbf{17.03} & 16.82 & 16.18 \\
NeuBoots & 17.94$\pm$0.74 & \textbf{14.98}$\pm$0.31 & \textbf{14.45}$\pm$0.31
\tabularnewline
\bottomrule
\end{tabular}
\caption{Rotated MNIST classification with different bootstrap sampling number $B$.}
\label{tab:rotatedMNIST}
\end{table*}

\paragraph{NeuBoots vs Dropout}

At first glance, NeuBoots is similar to Dropout 
in that the final neurons are multiplied by random variables.
However, random weights imposed by the Dropout are lack of connection to the loss function nor the working model, while the bootstrap weights of the NeuBoots appears in the loss function (\ref{eq:block-bootstrap-loss}) have explicit connections to the bootstrapping.
We briefly verify the effect of the loss function on the 3-layers MLP with the different number of hidden variables 50, 100, and 200 for the image classification task on MNIST datasets.
With batch normalization \cite{ioffe2015batch}, we have applied Dropout with  probability $p=0.1$ only to the final layer of MLP.
We measure the ECE, the negative log-likelihood (NLL), and the Brier score for comparisons.
NeuBoots and Dropout records same accuracy. 
However, Figure \ref{fig:NeuBoots-Dropout-MNIST} shows that the NeuBoots is more feasible for confidence-aware learning and clearly outperforms the Dropout in terms of ECE, NLL, and the Brier score. 

\begin{wrapfigure}{R}{0.35\textwidth}
\centering
    \includegraphics[width=0.35\textwidth]{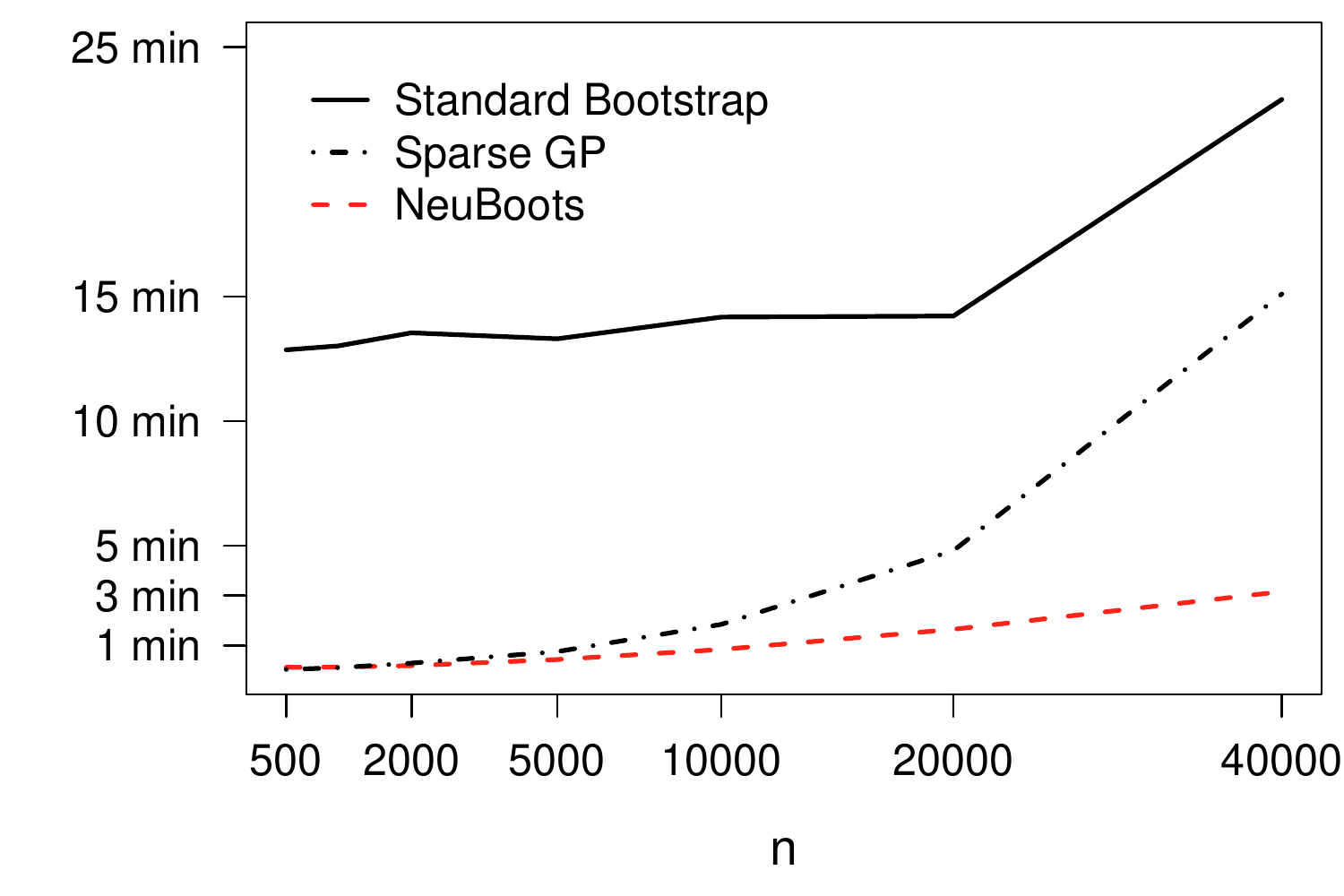}
    \caption{Comparison of computational time  with different numbers of training data $n$ 
    for the example in Figure \ref{fig:NeuBoots-GP-MCDropout}
    }\label{fig:time}
\end{wrapfigure}

\paragraph{Computation time and cost}

As we mentioned earlier, the algorithm evaluates the network from scratch for only once to store the tensor $F_{\theta_{2}}(X_{*})$, while the standard bootstrapping and MCDrop \cite{gal2016dropout} need repetitive feed-forward propagations.
To check this empirically, we measure the prediction time by ResNet-34 between NeuBoots and MCDrop on the test set of CIFAR-10 with Nvidia V100 GPUs.
NeuBoots predicts $B=100$ bootstrapping in 1.9s whereas MCDrop takes 112s to generate 100 outputs. 
Also, NeuBoots is computationally more efficient than the standard bootstrap and the sparse GP \cite{snelson2006sparse} (Figure \ref{fig:time}). 

\begin{table}[h]
    \centering
    \footnotesize
  \begin{tabular}{c | c c c}
    \toprule
    Method &  Training Time & Test Time & Memory Usage  
    \tabularnewline
    \midrule
    DeepEnsemble     & $O(LK)$ & $O(LK)$ & $O(MK+I)$ \\
    BatchEnsemble  & $O(LK)$  & $O(LK)$  & $O(M+IK)$\\
    MIMO  & $O(L + 2K)$  & $O(L+2K)$  & $O(M+IK)$\\
    NeuBoots & $O(L)$  & $O(L+K)$  & $O(M+I)$
    \tabularnewline
    \bottomrule
  \end{tabular}
    \caption{A comparison of computational costs. We use the following notations: $L$ the number of layers, $K$ the number of bootstrapping (or ensemble), $M$ the parameter size of a single model, $I$ memory size of input data.}
    \label{tab:comp_costs}
\end{table}

We also compare NeuBoots to MIMO \cite{havasi2021training} and BatchEnsemble \cite{wen2020batchensemble} in terms of training, test, and memory complexities (see Table \ref{tab:comp_costs}).
Since NeuBoots does not require repetitive forward computations, its training and test costs are $O(L)$ and $O(L+K)$, respectively, less than $O(L+2K)$ of MIMO and $O(LK)$ of BatchEnsemble.  
Note that MIMO needs to copy input images as many as $K$ to supply into input layers.
Even though it can compute in a single forward pass, it requires more memories to upload multiple inputs if the input data is high-dimensional (e.g., MRI/CT).
The memory complexity of BatchEnsemble is similar to the one of MIMO since the memory usage of fast weights in BatchEnsemble is proportional to the dimension of input and output.
This computational bottleneck is nothing to sneeze at in the application fields requiring on-device training or inference; however, the proposed method is free from such a problem since multiple computations occur only at the final layer.
For quantitative comparisons, we refer to Appendix \ref{sec:empirical_comparison_mimo_be}.

\paragraph{Diversity of predictions}

Diversity of predictions has been a reliable measure to examine over-fits and performance of uncertainty quantification for ensemble procedures \cite{rame2021dice,fort2020deep, wen2020batchensemble}. 
In the presence of overfitting, it is expected that the diversity of different ensemble predictions would be minimal because the resulting ensemble models would produce similar predictions that are over-fitted towards the training data points. 
To examine the diversity performance of NeuBoots, we consider various diversity measures including ratio-error, Q-statistics, correlation coefficient, and prediction disagreement (see \cite{aksela2003comparison, rame2021dice, fort2020deep}). 
For the CIFAR-100 along with DenseNet-100, Table \ref{tab:diversity} summarizes the results. 
NeuBoots outperforms MCDrop in every metrics of diversity.
Furthermore, NeuBoots shows comparable results with DeepEnsemble.

\begin{table}[h]
    \centering
    \footnotesize
    \begin{tabular}{c|c c c c}
    \toprule
      Method   & Ratio-error ($\uparrow$)  & Q-stat ($\downarrow$) & Correlation ($\downarrow$) & Disagreement ($\uparrow$)
      \tabularnewline
    \midrule
       DeepEnsemble & {\bf 98.00}  & {\bf 61.31} & 78.56 & 23.41 \\
      MCDrop & 27.38 & 96.33 & 92.00 & 10.40\\
        NeuBoots &  93.79 & 63.95 & {\bf 76.11} & {\bf 32.20} \\
      \bottomrule
    \end{tabular}
    \caption{A comparison of diversity performances.}
    \label{tab:diversity}
\end{table}

\section{Empirical Studies}
    \label{sec:emp}

In this section, we conduct the wide range of empirical studies of NeuBoots for uncertainty quantification and bagging performance. 
We apply NeuBoots to prediction calibration, active learning, out-of-distribution detection, bagging performance for semantic segmentation, and learning on imbalanced dataset with various deep convolutional neural networks.
Our code is open to the public\footnote{\url{https://github.com/sungbinlim/NeuBoots}}.

\subsection{Prediction Calibration}
    \label{sec:calibration}

\paragraph{Setting}
We evaluate the proposed method on the prediction calibration for image classification tasks. We apply  NeuBoots  to image classification tasks on CIFAR and SVHN with ResNet-110 and DenseNet-100.
We take $k=5$ predictions of MCDrop and DeepEnsemble for calibration. 
For fair comparisons, we set the number of bootstrap sampling $B=5$ as well, and fix the other hyperparameters same with baseline models.
All models are trained using SGD with a momentum of 0.9, an initial learning rate of 0.1, and a weight decay of 0.0005 with the mini-batch size of 128.
We use CosineAnnealing for the learning rate scheduler.
We implement MCDrop and evaluates its performance with dropout rate $p=0.2$, which is a close setting to the original paper.
For Deep Ensemble, we utilize adversarial training and the Brier loss function \cite{lakshminarayanan2017simple} and cross-entropy loss function \cite{Ashukha2020Pitfalls}.
For the metric, we evaluate the error rate, ECE, NLL, and Brier score. 
We also compute each method's training and prediction times to compare the relative speed based on the baseline.

\paragraph{Results}

\begin{figure}[ht]
    \centering
    \subfloat[ Reliability Diagram]{{\includegraphics[width=0.5\textwidth]{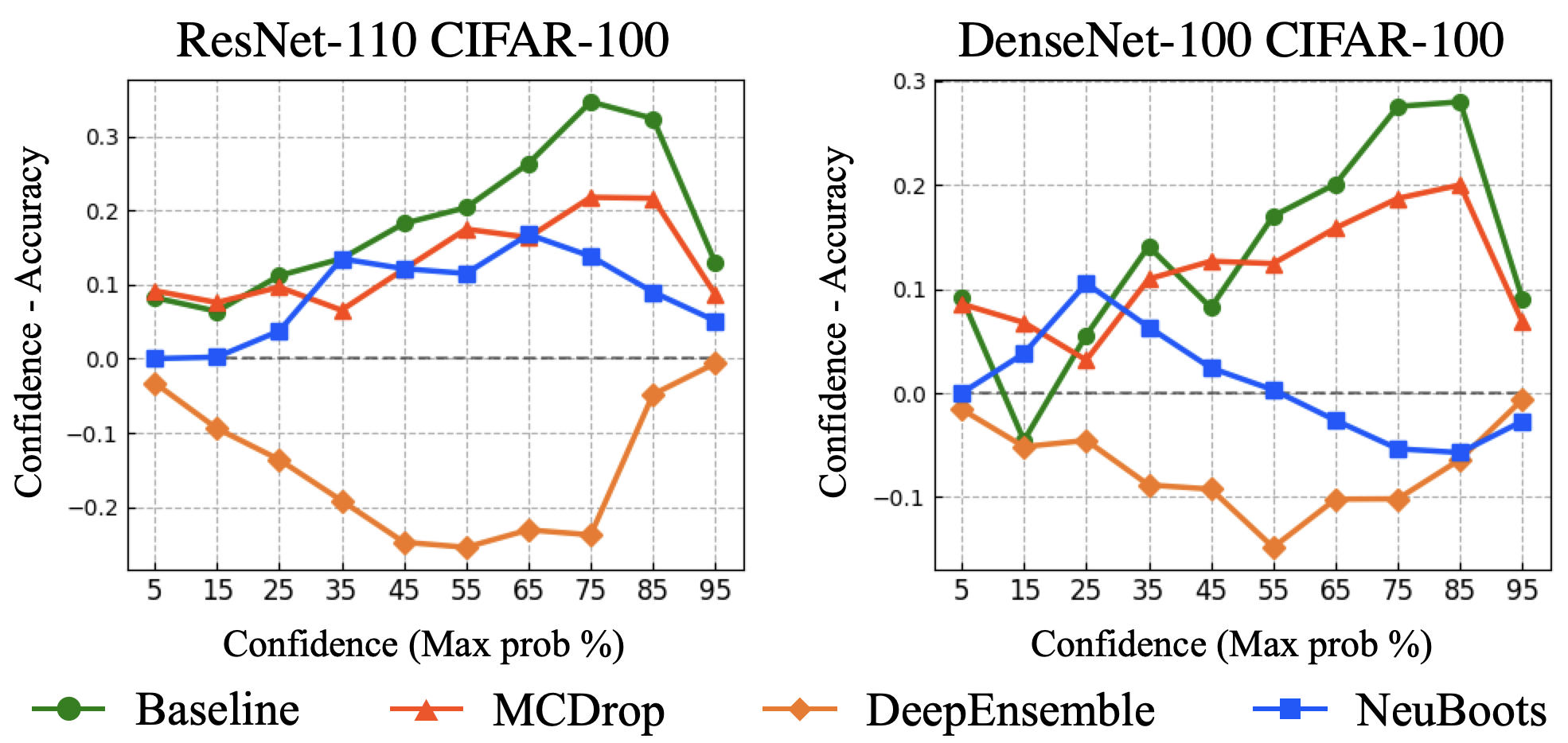} }}%
    \subfloat[ Active Learning]{{\includegraphics[width=0.5\textwidth]{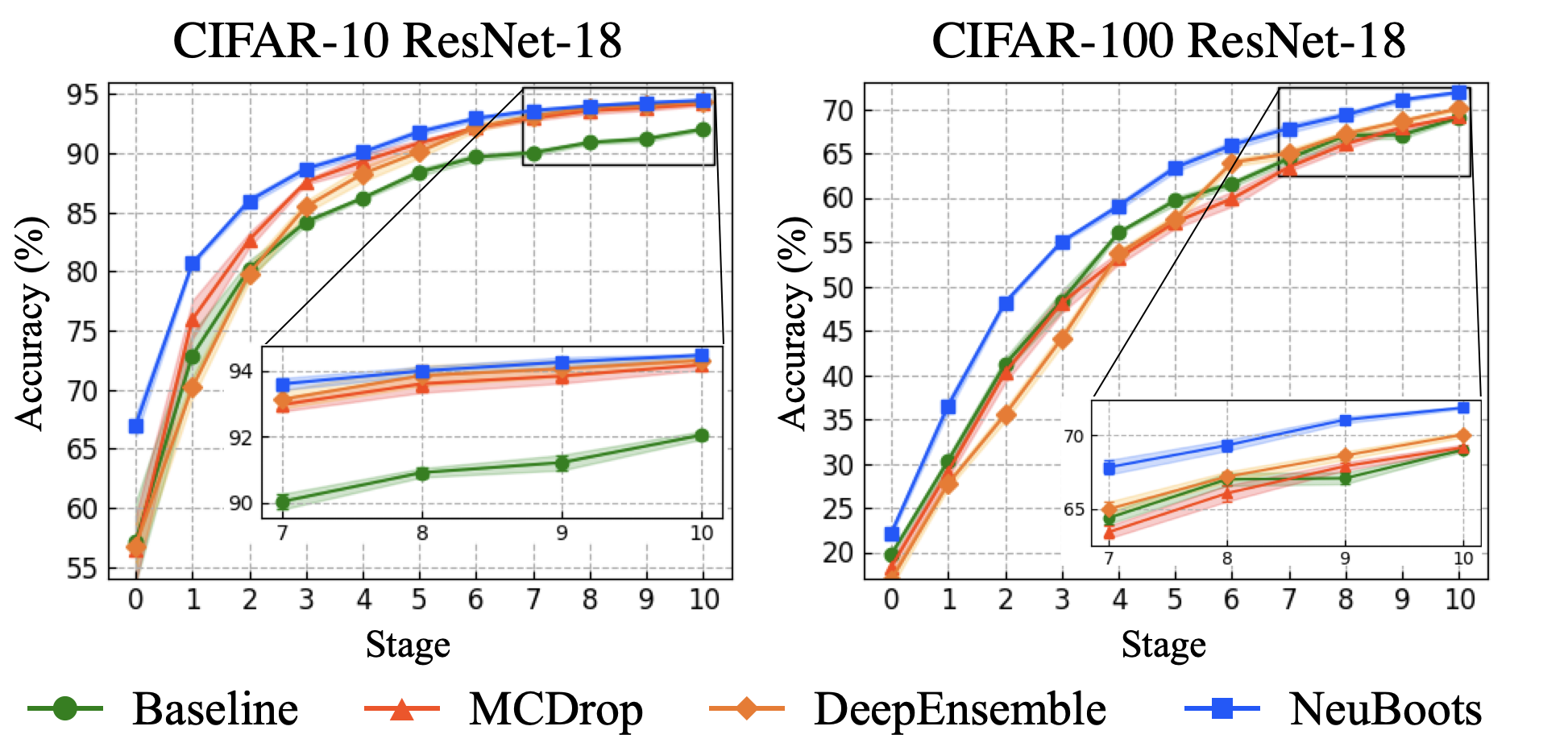} }}%
    \caption{(a) Comparison of reliability diagrams for ResNet-110 and DenseNet-100 on CIFAR-100. Confidence is the value of the maximal softmax output. A dashed black line represents a perfectly calibrated prediction. Points below this line indicate to under-confident predictions, whereas points above the line mean overconfident predictions. (b) Actice learning performance on CIFAR-10 (left) and CIFAR-100 (right) with Random, MCDrop, and NeuBoots. Curves are averaged over five runs and shaded regions denote the confidence intervals.}%
    \label{fig:calibration-active}
\end{figure}

See Table \ref{table_classification} and \ref{tab:table_classification_appendix} for empirical results. 
NeuBoots generally show a comparable calibration ability compared to MCDrop and DeepEnsemble.
Figure \ref{fig:calibration-active}.(a) shows the reliability diagrams of ResNet-110 and DenseNet-100 on CIFAR-100.
We observe that NeuBoots secures accuracy and prediction calibration in the image classification tasks with ResNet-110 and DenseNet-100.
NeuBoots is faster in prediction than MCDrop and DeepEnsemble at least three times.
Furthermore, NeuBoots shows faster in training than Deep Ensemble at least nine times.
This gap increases as the number of predictions $k$ increases.
It concludes that NeuBoots outperforms MCDrop and is comparable with DeepEnsemble in calibrating the prediction with the relatively faster prediction. 

\subsection{Active Learning}
    \label{sec:active}
    

\paragraph{Setting}

We evaluate the NeuBoots on the active learning with ResNet-18 architecture on CIFAR.
For a comparison, we consider MCDrop and DeepEnsemble with entropy-based sampling and random sampling.
We follow an ordinary process to evaluate the performance of active learning (see \cite{moon2020confidence}).
Initially, a randomly sampled 2,000 labeled images are given, and we train a model. 
Based on the uncertainty estimation of each model, we sample 2,000 additional images from the unlabeled dataset and add to the labeled dataset for the next stage.
We continue this process ten times for a single trial and repeat five trials for each model.

\paragraph{Results}
Figure \ref{fig:calibration-active}.(b) shows the sequential performance improvement on CIFAR-10 and CIFAR-100.
Note that CIFAR-100 is more challenging dataset than CIFAR-10.
Both plots demonstrate that NeuBoots is superior to the other sampling methods in the active learning task. 
NeuBoots records 71.6\% accuracy in CIFAR-100 and 2.5\% gap with MCDrop and DeepEnsemble.  
Through the experiment, we verify that NeuBoots has a significant advantage in active learning.


\subsection{Out-of-Distribution Detection}

\paragraph{Setting}

As an important application of uncertainty quantification, we have applied NeuBoots to detection of out-of-distribution (OOD) samples.
The setting for OOD is based on the Mahalanobis method \citep{lee2018simple}.
At first, we train ResNet-34 for the classification task only using the training set of the CIFAR-10 (in-distribution).
Then, we evaluate the performance of NeuBoots for OOD detection both in the test sets of in-distribution dataset and the SVHN (out-of-distribution).
Using a separate validation set from the testsets, we train a logistic regression based detector to discriminate OOD samples from in-distribution dataset.
For the input vectors of the OOD detector, we extract the following four statistics regarding logit vectors: the max of predictive mean vectors, the standard deviation of logit vectors, expected entropy, and predictive entropy, which can be computed by the sampled output vectors of NeuBoots.
To evaluate the performance of the detector, we measure the true negative rate (TNR) at 95\% true positive rate (TPR), the are under the receiver operating characteristic curve (AUROC), the area under the precision-recall curve (AUPR), and the detection accuracy. 
For comparison, we examine the baseline method \citep{hendrycks17baseline}, MCDrop, DeepEnsemble \cite{lakshminarayanan2017simple}, DeepEnsemble\_\text{CE} (trained with cross-entropy loss) \cite{Ashukha2020Pitfalls}, ODIN \citep{liang2018enhancing}, and Mahalanobis \citep{lee2018simple}. 

\paragraph{Results}

\begin{table*}[h] \center
\footnotesize
\begin{tabular}{c c c c c c}
\toprule
 \multirow{2}{*}{Method} & TNR & \multirow{2}{*}{AUROC} & Detection & AUPR & AUPR \\
  & at TPR 95\% &  & Accuracy & In & Out  
\tabularnewline
\midrule
 Baseline & 32.47 & 89.88 & 85.06 & 85.4 & 93.96 \\
 MCDrop & 51.4 & 92.01 & 89.46 & 86.82 & 95.41 \\
 DeepEnsemble \cite{lakshminarayanan2017simple} & 56.7 & 91.85 & 88.91 & 81.66 & 95.46 \\
 DeepEnsemble\_\text{CE} \cite{Ashukha2020Pitfalls} & 48.5 & 92.29 & 90.48 & 86.33 & 95.49 \\
 ODIN & 86.55 & 96.65 & 91.08 & 92.54 & 98.52 \\
 Mahalanobis & 54.51 & 93.92 & 89.13 & 91.54 & 98.52 \\
 Mahalanobis + Calibration & 96.42 & \textbf{99.14} & 95.75 & \textbf{98.26} & 99.6 \\
\midrule
 NeuBoots & 89.40 & 97.26 & 93.80 & 93.97 & 98.86 \\
 NeuBoots + Calibration & \textbf{99.00} & \textbf{99.14} & \textbf{96.52} & 97.78 & \textbf{99.68}
\tabularnewline
\bottomrule
\end{tabular}
\caption{OOD detection. 
All values are percantages and the best results are indicated in bold.}  
\label{table_OOD_cifar}
\end{table*}

Table \ref{table_OOD_cifar} shows NeuBoots significantly outperform the baseline method \cite{hendrycks17baseline}, DeepEnsemble \cite{lakshminarayanan2017simple, Ashukha2020Pitfalls}, and ODIN \citep{liang2018enhancing} without any calibration technique in OOD detection. 
Furthermore, with the input pre-processing technique studied in \cite{liang2018enhancing}, NeuBoots is superior to Mahalanobis \citep{lee2018simple} in most metrics, which employs both the feature ensemble and the input pre-processing for the calibration techniques. 
This validates NeuBoots can discriminate OOD samples effectively. 
In order to see the performance change of the OOD detector concerning the bootstrap sample size, we evaluate the predictive standard deviation estimated by the proposed method for different $B \in \{2, 5, 10, 20, 30\}$. 
Figure \ref{fig:ood-batch-size} illustrates that the NeuBoots successfully detects the in-distribution samples (top row) and the out-of-distribution samples (bottom row).



\subsection{Bagging Performance for Semantic Segmentation}
    \label{sec:segmentation}

\paragraph{Setting}

To demonstrate the applicability of NeuBoots to different computer vision tasks, we validate NeuBoots on PASCAL VOC 2012 semantic segmentation benchmark \cite{Everingham10} with DeepLab-v3 \cite{chen2017rethinking} based on the backbone architectures of ResNet-50 and ResNet-101.
We modify the final $1\times 1$ convolution layer after the Atrous Spatial Pyramid Pooling (ASPP) module by multiplying the channel-wise bootstrap weights.
This is a natural modification of the segmentation architecture analogous to the fully connected layer of the networks for classification tasks.
Additionally, we apply NeuBoots to real 3D image segmentation task on commercial ODT microscopy NIH3T3 \cite{Choi2021.05.23.445351} dataset, which is challenging for not only models but also human due to the 512 $\times$ 512 $\times$ 64 sized large resolution and endogenous cellular variability.
We use two different U-Net-like models for this 3D image segmentation task, which are U-ResNet and SCNAS.
We simply amend the bottleneck layer in the same way as the 2D version.
Same as an image classification task, we set $B=5$ and $k=5$.
For the remaining, we follow the usual setting.

\paragraph{Results}

\begin{table*}[!t]
\begin{adjustbox}{max width=\textwidth} \small
    \centering
    \begin{tabular}{c c c | c c c }
\toprule
 Dataset &
 Architecture & 
 Method & 
 mIoU$(\%)$ & 
 ECE$(\%)$ & 
 Relative Prediction Time
\tabularnewline
\midrule
 \multirow{10}{*}{$\underset{\text{(PASCAL VOC \cite{Everingham10})}}{\text{2D}}$} &
 \multirow{5}{*}{ResNet-50} & Baseline & 84.57$\pm$0.72 & 15.35$\pm$0.21 & 1.0 \\
 & & MCDrop & 87.81$\pm$1.83 & 6.6$\pm$0.1 & 5.4 \\
 & & DeepEnsemble \cite{lakshminarayanan2017simple} & 90.09$\pm$0.61 & 17.31$\pm$0.74 & 5.5 \\
 & & DeepEnsemble\_\text{CE} \cite{Ashukha2020Pitfalls} & 86.95$\pm$0.57 & 12.36$\pm$0.53 & 5.5 \\
 & & NeuBoots & \textbf{90.14}$\pm$2.17 & \textbf{6.00}$\pm$0.1 & \textbf{2.7} \\
 \cmidrule{2-6}
 & \multirow{5}{*}{ResNet-101} & Baseline & 85.35$\pm$0.23 & 15.49$\pm$0.44 & 1.0 \\
 & & MCDrop & 88.08$\pm$1.80 & 6.48$\pm$0.08 & 5.3 \\
 & & DeepEnsemble \cite{lakshminarayanan2017simple} & 90.40$\pm$0.11 & 17.94$\pm$0.03 & 5.3 \\
  & & DeepEnsemble\_\text{CE} \cite{Ashukha2020Pitfalls} & 87.48$\pm$0.09 & 11.52$\pm$0.02 & 5.3 \\
 & & NeuBoots & \textbf{90.56}$\pm$1.71 & \textbf{6.14}$\pm$0.11 & \textbf{2.5} \\
 \midrule
 \multirow{10}{*}{$\underset{\text{(NIH3T3 \cite{Choi2021.05.23.445351})}}{\text{3D}}$} & \multirow{5}{*}{U-ResNet} & Baseline & 61.54$\pm$1.14 & 1.85$\pm$0.19 & 1.0 \\
 & & MCDrop & 64.15$\pm$0.48 & 1.53$\pm$0.09 & 5.5 \\
 & & DeepEnsemble \cite{lakshminarayanan2017simple} & 59.71$\pm$1.82 & 1.78$\pm$0.29 & 5.5 \\
 & & DeepEnsemble\_\text{CE} \cite{Ashukha2020Pitfalls} & 65.71$\pm$1.69 & \textbf{0.94}$\pm$0.24 & 5.5 \\
 & & NeuBoots & \textbf{67.78}$\pm$1.01 & 1.67$\pm$0.19 & \textbf{3.5} \\
\cmidrule{2-6}
 & \multirow{5}{*}{SCNAS \cite{kim2019scalable}} & Baseline & 67.52$\pm$1.95 & 1.45$\pm$0.19 & 1.0 \\
 & & MCDrop & 65.37$\pm$1.13 & 0.64$\pm$0.17 & 5.2 \\
 & & DeepEnsemble \cite{lakshminarayanan2017simple} & 60.04$\pm$2.11 & 1.39$\pm$0.05 & 5.3 \\
 & & DeepEnsemble\_\text{CE} \cite{Ashukha2020Pitfalls} & 68.66$\pm$2.58 & 0.83$\pm$0.09 & 5.3 \\
 & & NeuBoots & \textbf{70.80}$\pm$1.58 & \textbf{0.63}$\pm$0.16 & \textbf{2.1} \\
\bottomrule
\end{tabular}
\end{adjustbox}
    \caption{Semantic segmentation. The best results are indicated in bold.}  
\label{table_segmentation_2d-3d}
\end{table*}

Table \ref{table_segmentation_2d-3d} shows NeuBoots significantly improves mean IoU and ECE compared to the baseline. 
Furthermore, similar to the image classification task, NeuBoots records faster prediction time than MCDrop and DeepEnsemble. 
This experiment indeed verifies that NeuBoots can be applied to the wider scope of computer vision tasks beyond image classification.

\subsection{Imbalanced Dataset}

\begin{figure}[ht]
    \centering
    \includegraphics[width=0.7\textwidth]{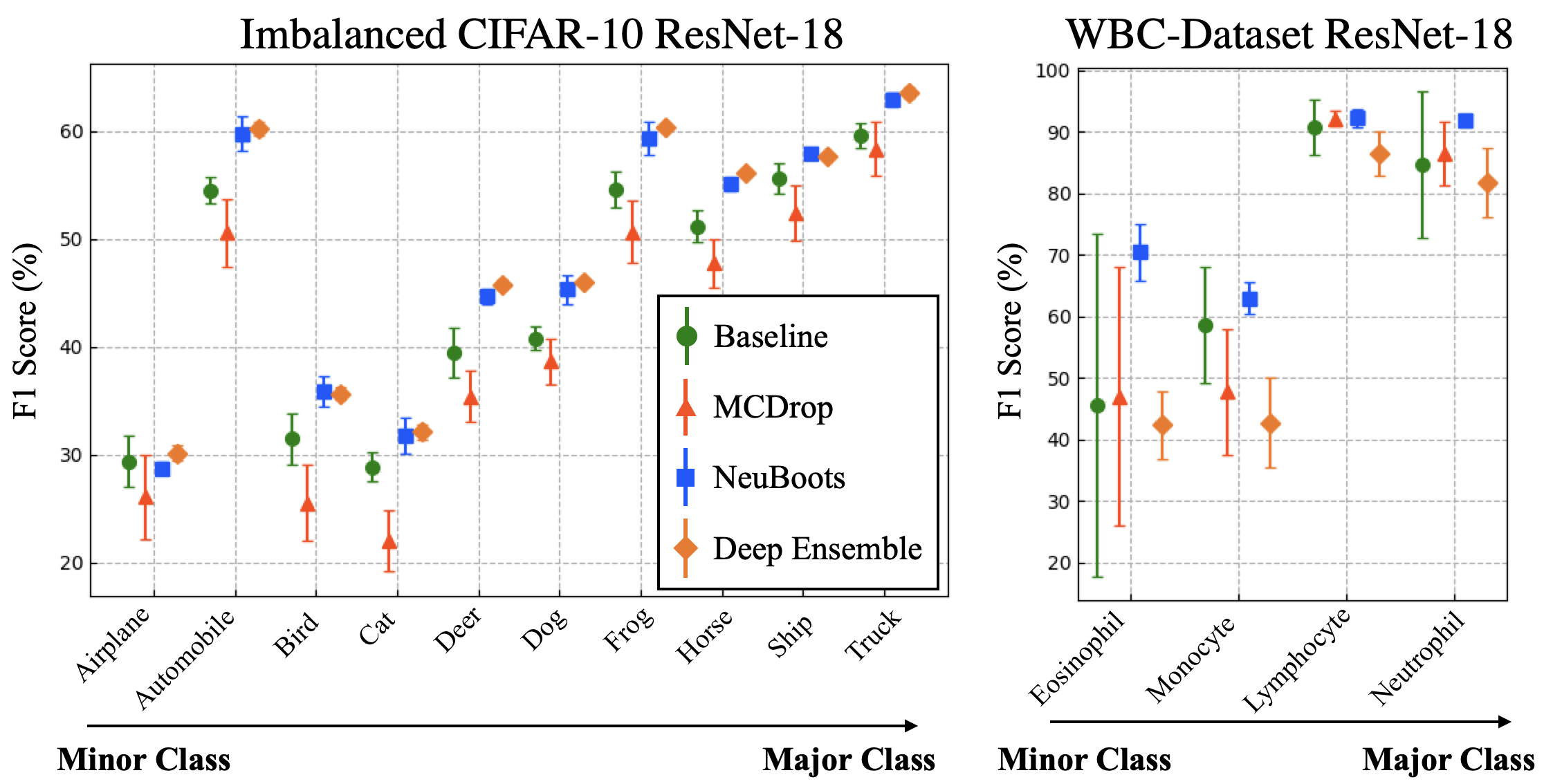}
    \caption{Comparisons of classification power for imbalance datasets. The minor class refers to the class with the least number of samples, and the major class refers to the highest number of samples.}
    \label{fig:imbalance}
\end{figure}

\paragraph{Setting}
To validate the efficacy for the imbalanced dataset, we have applied NeuBoots to two imbalance sets, the imbalanced CIFAR-10 and the white blood cell dataset with ResNet-18.
To conduct an imbalanced CIFAR-10, we randomly sampled from the training dataset of CIFAR-10 to follow a different distribution for each class, and the distribution is [{\it 50, 100, 150, 200, 250, 300, 350, 400, 450, 500}] for [{\it airplane, automobile, bird, cat, deer, dog, frog, horse, ship, truck}].
The white blood cell dataset was acquired using a commercial ODT microscopy, and each image is a grayscale of 80 $\times$ 80.
The dataset comprises four types of white blood cells, and the training distribution is [{\it 144, 281, 2195, 3177}] for [{\it eosinophil, monocyte, lymphocyte, neutrophil}].
ResNet-18 model and MCDrop extension are used as a baseline and comparative model with the same settings as Section \ref{sec:active}, respectively.
We measure the F1 Score for each class for evaluation.

\paragraph{Results}
Comparing the performance of Baseline, MCDrop, and DeepEmsenble, NeuBoots performs better on both imbalanced CIFAR-10 and the white blood cell dataset, as shown in Figure \ref{fig:imbalance}.
Especially, NeuBoots outperforms for eosinophil identification, the class with the lowest number of samples in the white blood cell dataset, with low variance.
This result shows that the NeuBoots boosts the prediction power for the fewer sampled classes with high stability via simple implementation.

\section{Related Work}


\paragraph{Bootstrapping Neural Network}
Since \cite{efron1979bootstrap} first proposed the nonparametric bootstrapping to quantify uncertainty in general settings, there has been  a rich amount of literature that investigate theoretical advantages of using bootstrap procedures for parametric models \citep{hall1986bootstrap, hall1992bootstrap, efron1987better}. 
For nerural networks, \cite{franke2000bootstrapping} investigated bootstrap consistency of one-layered MLP under some strong regularity conditions.
\cite{reed2014training} considered using a conventional nonparametric bootstrapping to robustify classifiers under noisy labeling. However, due to the nature of repetitive computations, its practical application to large-sized data sets is not trivial. 
\cite{nalisnick2017amortized} proposed an approximation of bootstrapping for neural network by applying amortized variational Bayes. Despite its computational efficiency, the armortized bootstrap does not induce the exact target bootstrap distribution, and its theoretical justification is lacking. 
Recently, \cite{lee2020bootstrapping} proposes a bootstrapping method for neural processes. 
They utilized residual bootstrap to resolve the data discard problem, but their approach is not scalable since it requires multiple encoder computations.


\paragraph{Ensemble Methods}
 Various advances of neural net ensembles have been made to improve computational efficiency and uncertainty quantification performance.
 \citet{havasi2021training} introduces Multiple Input Multiple Output (MIMO), that approximates independent neural nets by imposing multiple inputs and outputs, and \citet{wen2020batchensemble} proposes a low-rank approximation of ensemble networks, called BatchEnsemble. 
Latent Posterior Bayes NN (LP-BNN, \cite{franchi2020encoding}) extends the BatchEnsemble to a Bayesian paradigm imposing a VAE structure on the individual low-rank factors, and the LP-BNN outperforms the MIMO and the BatchEnsemble in prediction calibration and OOD detection, but its computational burden is heavier than that of the BatchEnsemble.   
 Stochastic Weight Averaging Gaussian (SWAG, \cite{maddox2020simple}) computes the posterior of the base neural net via a low-rank approximation with a batch sampling.  
 Even though these strategies reduces the computational cost to train each ensemble network, unlike NeuBoots, they still demand multiple optimizations, and its computational cost linearly increases as the ensemble size grows up.

\paragraph{Uncertainty Estimation}
There are numerous approaches to quantify the uncertainty in predictions of NNs. 
Deep Confidence \citep{cortes2018deep} proposes a framework to compute confidence intervals for individual predictions using snapshot ensembling and conformal prediction. 
Also, a calibration procedure to approximate a confidence interval is proposed based on Bayesain neural networks \citep{kuleshov2018accurate}.  
\citet{gal2016dropout} proposes MCDrop which captures model uncertainty casting dropout training in neural networks as an approximation of variational Bayes.
\citet{SmithGal2018Understanding} examines various measures of uncertainty for adversarial example detection. 
\citet{lakshminarayanan2017simple} proposes a non-Bayesian approach, called DeepEnsemble, to estimate predictive uncertainty based on ensembles and adversarial training. Compared to DeepEnsemble, NeuBoots does not require adversarial training nor learning multiple models. 

\section{Conclusion}
    \label{sec:conclusion}
We introduced a novel and scalable bootstrapping method, NeuBoots, for neural networks. 
We applied it to the wide range of machine learning tasks related to uncertainty quantification;
prediction calibration, active learning, out-of-distribution detection, and imbalanced datasets.
NeuBoots also demonstrates superior bagging performance over semantic segmentation.
Our empirical studies show that NeuBoots attains significant potential in quantifying uncertainty for large-sized applications, such as biomedical data analysis with high-resolution. 
As a future research, one can apply NeuBoots to natural language processing tasks using Transformor \cite{vaswani2017attention}.


\section{Acknowledgement}
The authors specially thanks to Dr. Hyokun Yun for his fruitful comments. Minsuk Shin would like to acknowledge support from the National Science
Foundation (NSF-DMS award \#2015528). 
This work was also supported by Institute of Information \& communications Technology Planning \& Evaluation(IITP) grant funded by the Korea government(MSIT)
(No.2020-0-01336, Artificial Intelligence Graduate School Program(UNIST)) and National Research Foundation of Korea(NRF) funded by the Korea government(MSIT)(2021R1C1C1009256).

\newpage
\section*{Checklist}


\begin{enumerate}

\item For all authors...
\begin{enumerate}
  \item Do the main claims made in the abstract and introduction accurately reflect the paper's contributions and scope?
    \answerYes{See Section \ref{sec:intro}}
  \item Did you describe the limitations of your work?
    \answerYes{See Section \ref{sec:conclusion}}
  \item Did you discuss any potential negative societal impacts of your work?
    \answerNo{We could not found any negative societal effect of our work.}
  \item Have you read the ethics review guidelines and ensured that your paper conforms to them?
    \answerYes{}
\end{enumerate}

\item If you are including theoretical results...
\begin{enumerate}
  \item Did you state the full set of assumptions of all theoretical results?
    \answerYes{See Appendix}
	\item Did you include complete proofs of all theoretical results?
    \answerYes{See Appendix}
\end{enumerate}

\item If you ran experiments...
\begin{enumerate}
  \item Did you include the code, data, and instructions needed to reproduce the main experimental results (either in the supplemental material or as a URL)?
    \answerYes{See Section \ref{sec:emp}}
  \item Did you specify all the training details (e.g., data splits, hyperparameters, how they were chosen)?
    \answerYes{}
	\item Did you report error bars (e.g., with respect to the random seed after running experiments multiple times)?
    \answerYes{We report error bars except for OOD experiment}
	\item Did you include the total amount of compute and the type of resources used (e.g., type of GPUs, internal cluster, or cloud provider)?
    \answerYes{See Section \ref{sec:regression}}
\end{enumerate}

\item If you are using existing assets (e.g., code, data, models) or curating/releasing new assets...
\begin{enumerate}
  \item If your work uses existing assets, did you cite the creators?
    \answerYes{}
  \item Did you mention the license of the assets?
    \answerYes{}
  \item Did you include any new assets either in the supplemental material or as a URL?
    \answerNo{NIH3T3 \cite{Choi2021.05.23.445351} is proprietary.}
  \item Did you discuss whether and how consent was obtained from people whose data you're using/curating?
    \answerNA{}
  \item Did you discuss whether the data you are using/curating contains personally identifiable information or offensive content?
    \answerNo{There is no issue on this problem. NTH3T3 \cite{Choi2021.05.23.445351} is cell-line data.}
\end{enumerate}

\item If you used crowdsourcing or conducted research with human subjects...
\begin{enumerate}
  \item Did you include the full text of instructions given to participants and screenshots, if applicable?
    \answerNA{Unnecessary}
  \item Did you describe any potential participant risks, with links to Institutional Review Board (IRB) approvals, if applicable?
    \answerNA{Unnecessary}
  \item Did you include the estimated hourly wage paid to participants and the total amount spent on participant compensation?
    \answerNA{Unnecessary}
\end{enumerate}

\end{enumerate}


\newpage

\clearpage
\newpage

\appendix

\onecolumn
\title{Neural Bootstrapper: Supplementray Material}

\section{Block Bootstrapping}
\label{appendix-sec:block-boots}

\subsection*{Implementation}

Let $I_{1},\ldots, I_{S}$ denotes the index sets of exclusive $S$ blocks where $S \ll n$. 
We allocate the index of data $[n]$ to each block $I_{1},\ldots, I_{S}$ by the stratified sampling to balance among classes. 
Let index function $u:[n] \to [S]$ denotes such assignment so $u(i)=s$ if $i\in I_{s}$.
Then, given some weight distribution $H_{\boldsymbol{\alpha}}$ on $\mathcal{W}_{S}\subset\mathbb{R}_{+}^{S}$, we impose the same value of weight on all elements in a block such as,
$w_{i} = \alpha_{u(i)}$ for  $i\in [n]$,
where $\bm\alpha=\{\alpha_{1},\dots,\alpha_{S} \} \sim H_{\boldsymbol{\alpha}}$.
We write an allocated weight vector as $\mathbf{w}_{\bm\alpha}=\{ \alpha_{u(1)},\dots,\alpha_{u(n)} \}\in\mathcal{W}_{n}$. 
Similar with GBS, setting $H_{\boldsymbol{\alpha}}=S\times \text{Dirichlet}(1,\dots,1)$ induces a block version of the RWB, and imposing $H_{\boldsymbol{\alpha}}= \text{Multinomial}(S;1/S,\dots,1/S)$ results in a block nonparametric bootstrap.
We remark that the Dirichlet distribution with a uniform parameter of one can be easily approximated by independent exponential distribution. 
That is, $z_i/\sum_{k=1}^n z_k\sim \mbox{Dirichlet}( 1,\dots,1)$ for independent and identically distributed $z_i\sim \text{Exp}(1)$. 
Due to the fact that $\sum_{i=k}^nz_k/n \approx 1$ by the law of large number for a moderately large $n$, $n^{-1}\times\{z_1,\dots,z_n\}$ approximately follows the Dirichlet distribution. 
This property is convenient in a sense that we do not need to consider the dependence structure in $\mathbf{w}$, and simply generate independent samples from $\text{Exp}(1)$ to sample the bootstrap weight. 
We use this block bootstrap as a default of the NeuBoots in sequel. 
The proposed procedure asymptotically converges towards the same target distribution where the conventional non-block bootstrap converges to, and under some mild regularity conditions.
Theoretically, the block bootstrap asymptotically approximates the non-blocked bootstrap well as the number of blocks $S$ increases as $n\to\infty$ (see Theorem \ref{theo:Blockbootstrap}).

\subsection*{Asymptotics of Block Bootstrap}
We shall rigorously investigate asymptotic equivalence between the blocked bootstrap and the non-blocked bootstrap. To ease the explanation for theory, we introduce some notation here. We distinguish a random variable $Y_i$ and its observed value $y_i$, and we assume that the feature $X_1,X_2,\dots$ is deterministic. the Euclidean norm is denoted by $\norm{\cdot}$, and the norm of a $L_2$ space is denoted by $\norm{\cdot}_2$. Also, to emphasize that the bootstrap weight $\bw$ depends on $n$, we use $\bw_n$. Let $Y_1,Y_2,\dots$ be i.i.d. random variables from the probability measure space $(\Omega, \mathcal{F}, \mathbb{P}_0)$. We denote the empirical probability measure by $\widehat{\mathbb P}_n:=\sum_{i=1}^n\delta_{Y_i}/n$, where $\delta_{x}$ is  a discrete point mass at $x\in\mathbb{R}$, and let $\mathbb{P}g = \int g d\mathbb{P}$, where $\mathbb{P}$ is a probability measure and $g$ is a  $\mathbb{P}$-measurable function. 
Suppose that $\sqrt{n}(\widehat {\mathbb{P}}_n-\mathbb{P}_0)$ weakly converges to a probability measure $\mathbb{T}$ defined on some sample space and its sigma field $(\Omega',\mathcal{F}')$. In the regime of bootstrap, what we are interested in is to estimate $\mathbb{T}$ by using some weighted  empirical distribution that is $\widehat{\mathbb P}_n^* = \sum_{i=1}^n w_{i}\delta_{Y_i}$, where $w_1, w_2, \dots$ is an i.i.d. weight random variable from a probability measure $\mathbb{P}_{\mathbf{w}}$. In the same sense, the probability measure acts on the block bootstrap is denoted by $\mathbb{P}_{\bw_{\bm\alpha}}$. We state a primary condition on bootstrap theory as follows:  
\begin{eqnarray}\label{eq:wconvergecond} 
\sqrt{n}(\widehat{\mathbb{P}}_ng - \mathbb{P}_0g) \to \mathbb{T}g\:\:\:\mbox{for }g\in\mathcal{D}\:\:\:\mbox{and}\:\:\:\mathbb{P}_0g_{\mathcal{D}}^{2}<\infty, 
\end{eqnarray}
 where  $\mathcal{D}$ is  a collection of some continuous functions of interest, and  $g_{\mathcal{D}}(\omega) = \sup_{g\in\mathcal{D}}|g(\omega)|$ is the envelope function on $\mathcal{D}$. This condition means that there exists a target probability measure and the functions of interest should be square-bounded. 
 
 Based on this condition, 
the following theorem states that  the block bootstrap asymptotically induces the same bootstrap distribution with that of non-block bootstrap. All proofs of theorems are deferred to the supplementary material.
\begin{thm}
    \label{theo:Blockbootstrap}
Suppose that \eqref{eq:wconvergecond} holds and $\{\alpha_1,\dots,\alpha_S\}^\T\sim S\times \text{Dirchlet}(1,\dots,1)$ with $w_{i} = \alpha_{u(i)}$. 
We assume some regularity conditions introduced in the supplementary material, and also assume  $S\to\infty$ as  $n\to\infty$. Then, for a $r_n$ such that $\norm{\widehat{f}-f_0}_2=O_{\mathbb{P}_{\mathbf{w}}}(\zeta_nr_n^{-1})$ for any diverging sequence $\zeta_n$, 
 \begin{eqnarray}\label{eq:unifconvg}
\sup_{x\in\mathcal{X},U\in\mathcal{B}} \left\lvert{\mathbb{P}_{{\mathbf w}}\left\{r_n (\widehat f_{{\mathbf w}}(x)-\widehat f(x)) \in U\right\}-\mathbb{P}_{\bw_{\bm\alpha}}\left\{ r_n(\widehat f_{\bw_{\bm{\alpha}}}(x)-\widehat f(x)) \in U\right\}}\right\rvert\to0,
 \end{eqnarray}
in $\mathbb{P}_0$-probability, where  $\mathcal{B}$ is the Borel sigma algebra.
 \end{thm}
Recall that the notation is introduced in Section \ref{sec:GBS}. 
\cite{praestgaard1993exchangeably} showed that the following conditions on the weight distribution to derive bootstrap consistency for general settings:

\noindent $\mathbf{W1}$. $\bw_n$ is exchangeable for $n=1,2,\dots$.\\
\noindent $\mathbf{W2}$. $w_{n,i}\geq 0$ and $\sum_{i=1}^nw_{n,i}=n$ for all $n$.\\
\noindent $\mathbf{W3}$. $\sup_n\norm{w_{n,1}}_{2,1}<\infty$, where $\norm{w_{n,1}}_{2,1}=\int\sqrt{\mathbb{P}_{\mathbf{w}}(w_{n,1}\geq t)}dt$. \\
\noindent $\mathbf{W4}$. $\lim_{\lambda\to\infty}\limsup_{n\to\infty}\sup_{t\geq\lambda}t^2\mathbb{P}_{\mathbf{w}}(w_{n,1}\geq t) = 0$.\\
\noindent $\mathbf{W5}$. $n^{-1}\sum_{i=1}^n(w_{n,i}-1)^2\to 1$ in probability.
 
 Under {\bf W1}-{\bf W5}, combined with \eqref{eq:wconvergecond}, showed that  $\sqrt{n}(\widehat{\mathbb{P}}_n^*-\widehat{\mathbb{P}}_n)$ weakly converges to $\mathbb{T}$. It was proven that the Dirichlet weight distribution satisfies {\bf W1}-{\bf W5}, and we first show that the Dirichlet weight distribution for the blocks  also satisfies the condition. Then, the block bootstrap of the empirical process is also consistent when the classical bootstrap of the empirical process is consistent.  
 
 
  Since the block bootstrap randomly assigns subgroups, the distribution of $\bw_n$ is exchangeable, so the condition {\bf W1} is satisfied. The condition  $\bf W2$ and $\bf W3$ are trivial. Since a Dirichlet distribution with a unit constant parameter can be approximated by a pair of independent exponential random variables; i.e $\{z_1/\sum_{i=1}^S z_i,\dots, z_S/\sum_{i=1}^S z_i\}\sim Dir(1,\dots,1)$, where $z_i\overset{i.i.d.}\sim \exp(1)$. Therefore, $S\times Dir(1,\dots,1)\approx \{z_1,\dots,z_S\}$, if $S$ is large enough. This fact shows that $t^2\mathbb{P}_{\mathbf{w}}(w_{n,1}\geq t)\approx t^2\mathbb{P}_z(z_1\geq t)$, and it follows that $\mathbb{P}_z(z_1\geq t) = \exp(-t)$, so {\bf W4} is shown. The condition {\bf W5} is trivial by the law of large number. Then, under {\bf W1}-{\bf W5}, Theorem 2.1 in \cite{praestgaard1993exchangeably} proves that 
\begin{align}\label{eq:BTconv}
\sqrt{n}(\widehat{\mathbb{P}}_n^*-\widehat{\mathbb{P}}_n)\Rightarrow \mathbb{T},
\end{align}  
  where the convergence ``$\Rightarrow$'' indicates weakly convergence.


We denote the true neural net parameter by $\phi_0$  such that $f_0 = f_{\phi_0}$, where $f_0$ is the true function that involves in the data generating process, and $\widehat \phi$ and $\widehat \phi_{\mathbf{w}}$ are  the minimizers of the \eqref{eq:BTfunctional} for one-vector (i.e. $\mathbf{w}=(1,\ldots, 1)$) and given $\mathbf{w}$, respectively.  This indicates that $\widehat f = f_{\widehat\phi}$ and $\widehat f_{\mathbf{w}} = f_{\widehat\phi_{\mathbf{w}}}$. Then, our objective function can be expressed as minimizing $\widehat P_n L(f_\phi(X),y)$ with respect to $\phi$. We further assume that 

\noindent {\bf A1. } the true function belongs to the class of neural network, i.e. $f_0\in\mathcal{F}$. 

\noindent {\bf A2. }$
\sup_{x\in\mathcal{X},U\in\mathcal{B}} \left\lvert{\mathbb{P}_{{\mathbf w}}\left\{r_n (\widehat f_{{\mathbf w}}(x)-\widehat f(x)) \in U\right\}-\mathbb{P}_{0}\left\{ r_n(\widehat f(x)- f_0(x)) \in U\right\}}\right\rvert\to0,
$

in $\mathbb{P}_0$-probability, where $f_0$ is the true function that involves in the data generating process.

\noindent {\bf A3. }Suppose that $\sum_{i=1}^n\frac{\partial}{\partial\phi} L(f_{\widehat\phi}(X_i),y_i)=0$, $\sum_{i=1}^n \frac{\partial}{\partial\phi}w_iL(f_{\widehat\phi_{\mathbf{w}}}(X_i),y_i)=0$ for any $\bw$, and $ \mathbb{E}_0[\frac{\partial}{\partial\phi}L(f_{\phi_0}(X),y)]  = 0 $.  

\noindent {\bf A4. } $\mathcal H$ is in $\mathbb{P}_0$-Donsker family, where $\mathcal{H}=\{\frac{\partial}{\partial\phi}L(f_\phi(\cdot),\cdot): \phi\in \Phi\}$; i.e. $\sqrt{n}(\widehat{\mathbb{P}}_ng - \widehat{\mathbb{P}}_0g) \to  \mathbb{T}g$ for $g\in \mathcal{H}$ and $\mathbb{P}_0 g^2_{\mathcal H} < \infty$.

These conditions assume that the classical weighted bootstrap is consistent, and a rigorous theoretical investigation of this consistency is non-existent at the current moment.
However, we remark that the main purpose of this theorem is to show that the considered block bootstrap induces asymptotically the same result from the classical non-block bootstrap so that the use of the block bootstrap is at least asymptotically equivalent to the classical counterpart. 
In this sense, it is reasonable to assume that the classical bootstrap is consistent.   

Then, it follows that 
\begin{eqnarray*}
&&\sup_{x\in\mathcal{X},U\in\mathcal{B}} \left\lvert{\mathbb{P}_{{\mathbf w}}\left\{r_n (\widehat f_{{\mathbf w}}(x)-\widehat f(x)) \in U\right\}-\mathbb{P}_{{\mathbf w}_\alpha}\left\{ r_n(\widehat f_{{\mathbf w_{\mathbf{\alpha}}}}(x)-\widehat f(x)) \in U\right\}}\right\rvert\\
&\leq& \sup_{x\in\mathcal{X},U\in\mathcal{B}} \left\lvert{\mathbb{P}_{{\mathbf w}}\left\{r_n (\widehat f_{{\mathbf w}}(x)-\widehat f(x)) \in U\right\}-\mathbb{P}_{0}\left\{ r_n(\widehat f(x)- f_0(x)) \in U\right\}}\right\rvert \\
&&+\sup_{x\in\mathcal{X},U\in\mathcal{B}} \left\lvert{\mathbb{P}_{{\mathbf w}_{\mathbf \alpha}}\left\{r_n (\widehat f_{{\mathbf w}_{\mathbf \alpha}}(x)-\widehat f(x)) \in U\right\}-\mathbb{P}_{0}\left\{ r_n(\widehat f(x)- f_0(x)) \in U\right\}}\right\rvert.
\end{eqnarray*}
The first part in the right-hand side of the inequality converges to 0 by {\bf A1}. Also, the second part also converges to 0. That is because  the empirical process of the block weighted bootstrap is asymptotically equivalent to the classical RWB, so {\bf A2} and {\bf A3} guarantees that the asymptotic behavior of the bootstrap solution should be consistent as the classical counterpart does. \qed

\section{Additional Experimental Results}



\subsection{NeuBoots vs Amortized Bootstrapping}


\begin{figure}[h]
\centering
    \includegraphics[width=\textwidth]{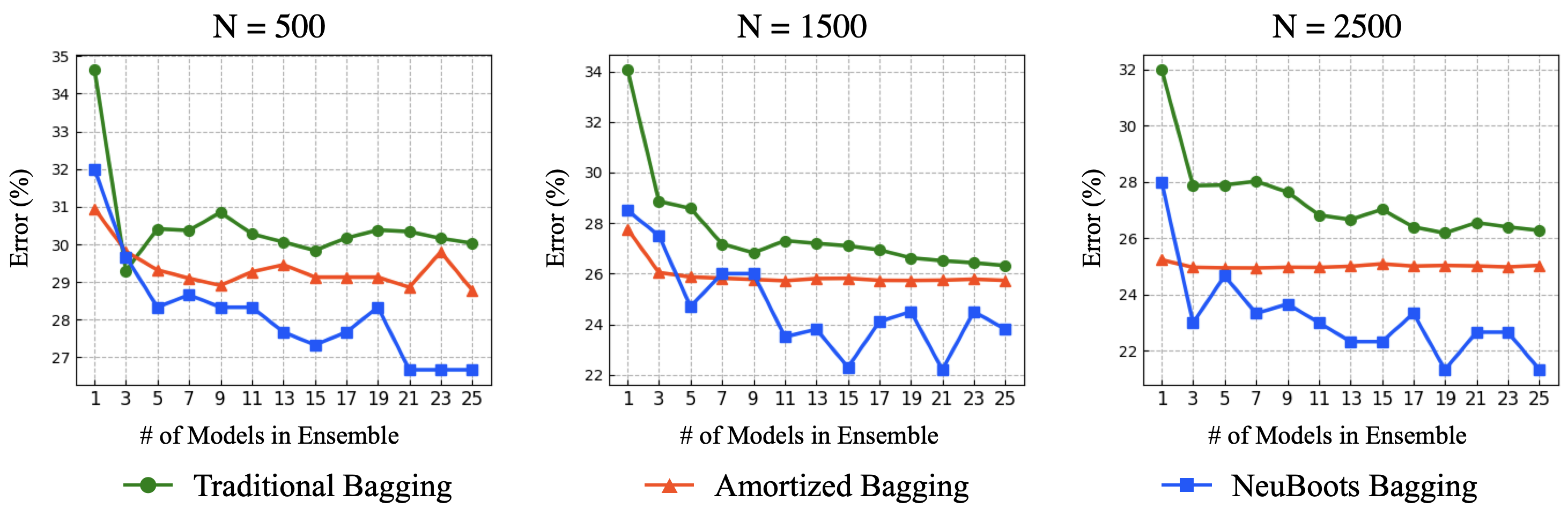}
    \caption{Comparison between standard bootstrapping, amortized bootstrapping \cite{nalisnick2017amortized}, and NeuBoots in Classification.}
    \label{fig:amortized-bootstrap-classification}
\end{figure}

\begin{figure}[h]
\centering
    \includegraphics[width=0.75\textwidth]{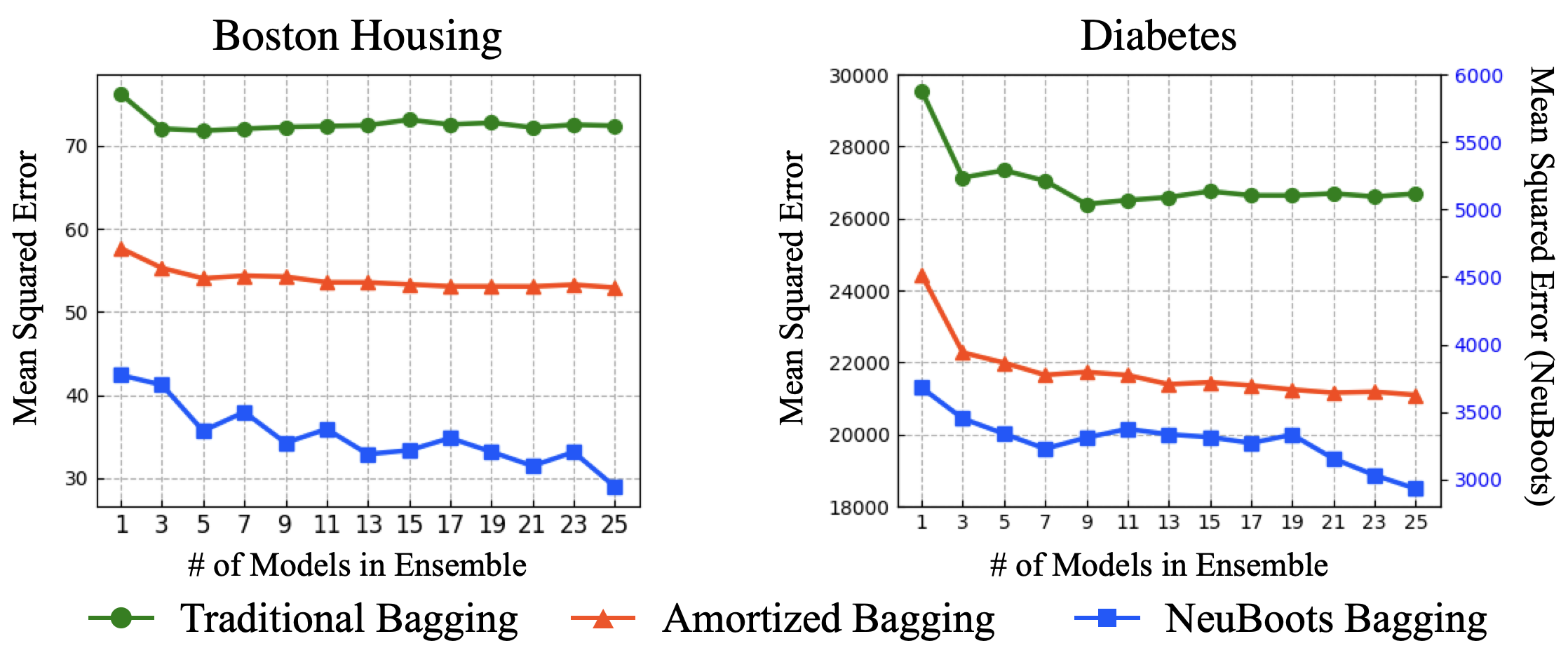}
    \caption{Comparison between standard bootstrapping, amortized bootstrapping\cite{nalisnick2017amortized}, and NeuBoots in Regression.}
    \label{fig:amortized-bootstrap-regression}
\end{figure}

\subsection{Computation time and costs}
\label{sec:empirical_comparison_mimo_be}

We supplement the comparison between MIMO, BatchEnsemble, and NeuBoots. 
We estimate training time and prediction during one epoch and measure the memory cost of those methods in ResNet-110 on CIFAR-100.

\begin{table}[h]
    \centering
    \footnotesize
  \begin{tabular}{c | c c c}
    \toprule
    Method &  Training Time (sec) & Test Time (sec) & Memory Usage (mb)
    \tabularnewline
    \midrule
    Baseline & 29.1 & 1.69 & 1605 \\
    MCDrop  & 29.2 & 8.45 & 1605 \\
    DeepEnsemble & 145.5 & 8.45 & 7600 \\
    BatchEnsemble & 217.1  & 6.59  & 2474 \\
    MIMO  & 30.2  & 1.87  & 1605 \\
    NeuBoots & 29.2  & 1.71  & 1605
    \tabularnewline
    \bottomrule
  \end{tabular}
    \caption{A comparison of training time, test time, and memory cost. }
    \label{tab:comp_cost_1}
\end{table}

We verified that NeuBoots benefits computational efficiency compared to BatchEnsemble. 
Interestingly, the above experiment shows that the computational cost of MIMO is similar to that of NeuBoots\footnote{The memory comparison between NeuBoots and MIMO seems unclear due to the small resolution of CIFAR. We observed NeuBoots has lower memory cost than MIMO if we change the model from ResNet to DenseNet.}. 
This result is because we used a convolutional neural network that benefits channel parallelism thanks to GPU. 
However, if we apply MIMO to MLP, the result is different: for 5-layer MLP ($K=3$), NeuBoots takes 7.57 seconds for one epoch training, MIMO takes 41.09 seconds, as we expected. 
Also, we observed that growing the number of the ensemble makes the proportion gap increase: for $K=5$, NeuBoots takes 7.68 seconds, MIMO takes 62.47 seconds. 
Furthermore, the below table shows that MIMO requires more memory cost in test time as the number of ensembles increases, as we expected in Table \ref{tab:comp_costs}.

\begin{table}[h]
    \centering
    \footnotesize
  \begin{tabular}{c | c c c c c}
    \toprule
    Method &  $K=1$ & $K=2$ & $K=3$ & $K=4$ & $K=5$
    \tabularnewline
    \midrule
    MIMO  & 1515  & 1633 & 1741 & 1885 & 2067  \\
    NeuBoots & 1515  & 1517  & 1517 & 1517 & 1517
    \tabularnewline
    \bottomrule
  \end{tabular}
    \caption{A comparison between MIMO and NeuBoots in memory cost (mb) as the number of ensemble increases. }
    \label{tab:comp_cost_2}
\end{table}

\subsection{Prediction Calibration}

\begin{table*}[!h] \center
\tiny
\begin{tabular}{c c | c c c c c c c}
\toprule
 \multirow{2}{*}{Architecture} & \multirow{2}{*}{Method} & Relative & Relative & \multirow{2}{*}{Error Rate$(\%)$} & \multirow{2}{*}{ECE$(\%)$} & \multirow{2}{*}{NLL$(\%)$} & \multirow{2}{*}{Brier Score$(\%)$} \\
 & & Training Time & Prediction Time & & & &
\tabularnewline
\midrule
  \multirow{4}{*}{ResNet-110}  & Baseline & 1.0 & 1.0 & 26.69$\pm$0.35 & 16.43$\pm$0.15 & 14.19$\pm$0.71 & 42.09$\pm$0.51 \\
  & MCDrop & 1.1 & 5.0 & 26.45 $\pm$0.08 & 13.65$\pm$1.25 & 13.16$\pm$0.64 & 40.46$\pm$0.30 \\
  & DeepEnsemble \cite{lakshminarayanan2017simple} & 9.5 & 5.0 & 34.84$\pm$0.21 & 27.33$\pm$4.92 & 18.69$\pm$1.44 & 56.12$\pm$3.02 \\
  & DeepEnsemble \cite{Ashukha2020Pitfalls} & 5.0 & 5.0 & \textbf{24.28}$\pm$0.11 & \textbf{4.74}$\pm$0.17 & \textbf{7.05}$\pm$0.28 & \textbf{28.29}$\pm$0.12 \\
  & NeuBoots & \textbf{1.1} & \textbf{1.2} & 26.53$\pm$0.19 & 8.13$\pm$0.28 & 15.68$\pm$0.31 & 39.31$\pm$0.64 \\
 \midrule
 \multirow{4}{*}{DenseNet-100}  & Baseline & 1.0 & 1.0 & 24.02$\pm$0.3 & 12.38$\pm$0.21 & 10.93$\pm$0.34 & 36.40$\pm$0.63 \\
 & MCDrop & 1.1 & 5.0 & 23.88$\pm$0.09 & 9.49$\pm$0.35 & 10.22$\pm$0.86 & 34.94$\pm$0.67 \\
 & DeepEnsemble \cite{lakshminarayanan2017simple} & 9.5 & 5.0 & 25.51$\pm$0.24 & 6.67$\pm$5.06 & 9.66$\pm$0.24 & 35.33$\pm$1.21 \\
 & DeepEnsemble \cite{Ashukha2020Pitfalls} & 5.0 & 5.0 & \textbf{20.16}$\pm$0.21 & 4.74$\pm$0.42 & \textbf{7.07}$\pm$0.14 & \textbf{30.29}$\pm$0.12 \\
 & NeuBoots & \textbf{1.1} & \textbf{1.3} & 23.46$\pm$0.09 & \textbf{2.38}$\pm$0.12 & 11.58$\pm$0.13 & 34.67$\pm$0.24 
 \tabularnewline
\bottomrule
\end{tabular}
\caption{Comparison of the relative training speed, relative prediction speed, error rate, ECE, NLL, and Brier on CIFAR-100. For each metric, the lower value means the better. Relative training and relative prediction times are normalized with respect to the baseline method. Best results are indicated in bold. 
}  
\label{table_classification}
\end{table*}

\begin{table*}[!h] \center
\tiny
\begin{tabular}{c | c c | c c c c c c c}
\toprule
 \multirow{2}{*}{Data} & \multirow{2}{*}{Architecture} & \multirow{2}{*}{Method} & Relative & Relative & \multirow{2}{*}{Error Rate$(\%)$} & \multirow{2}{*}{ECE$(\%)$} & \multirow{2}{*}{NLL$(\%)$} & \multirow{2}{*}{Brier Score$(\%)$} \\
 & & & Training Time & Prediction Time & & & &
\tabularnewline
\midrule
 \multirow{8}{*}{CIFAR-10} & \multirow{4}{*}{ResNet-110}  & Baseline & 1.0 & 1.0 & 5.89 & 4.46 & 3.34 & 10.2 \\
 & & MCDrop & 1.0 & 5.0 & 5.93 & 3.96 & 2.57 & 9.7 \\
 & & DeepEnsemble & 9.5 & 5.0 & \textbf{5.44} & 5.72 & \textbf{2.43} & \textbf{8.81} \\
 & & NeuBoots & \textbf{1.1} & \textbf{1.2} & 5.65 & \textbf{0.89} & 3.28 & 9.32 \\
 \cmidrule{2-9}
 & \multirow{4}{*}{DenseNet-100}  & Baseline & 1.0 & 1.0 & 5.13 & 3.2 & 2.23 & 8.3 \\
 & & MCDrop & 1.1 & 5.0 & 4.95 & 2.72 & 1.93 & 8.1 \\
 & & DeepEnsemble & 9.5 & 5.0 & 4.63 & \textbf{0.54} & \textbf{1.46} & \textbf{6.74} \\
 & & NeuBoots & \textbf{1.1} & \textbf{1.3} & \textbf{4.0} & 2.87 & 2.82 & 8.66 \\
\midrule
  \multirow{8}{*}{SVHN} 
 & \multirow{4}{*}{ResNet-110}  & Baseline & 1.0 & 1.0 & 3.55 & 2.39 & 1.75 & 5.8 \\
 & & MCDrop & 1.1 & 5.0 & 3.64 & 1.8 & 1.73 & 6.11 \\
 & & DeepEnsemble & 9.5 & 5.0 & \textbf{2.65} & 1.78 & \textbf{1.2} & \textbf{4.16} \\
 & & NeuBoots & \textbf{1.1} & \textbf{1.2} & 3.51 & \textbf{0.96} & 1.48 & 5.6 \\
 \cmidrule{2-9}
 & \multirow{4}{*}{DenseNet-100}  & Baseline & 1.0 & 1.0 & 3.6 & 3.2 & 2.23 & 8.3 \\
 & & MCDrop & 1.1 & 5.0 & 3.6 & 1.6 & 1.62 & 5.89 \\
 & & DeepEnsemble & 9.5 & 5.0 & \textbf{2.68} & 1.55 & \textbf{1.18} & \textbf{4.23} \\
 & & NeuBoots & \textbf{1.1} & \textbf{1.3} & 3.65 & \textbf{0.47} & 1.49 & 5.7 
 \tabularnewline
\bottomrule
\end{tabular}
\caption{Comparison of the relative training speed, relative prediction speed, error rate, ECE, NLL, and Brier on various datasets and architectures. For each metric, the lower value means the better. Relative training and relative prediction times are normalized with respect to the baseline method. Best results are indicated in bold. 
}  
\label{tab:table_classification_appendix}
\end{table*}

\newpage

\subsection{Dropout vs NeuBoots}

\begin{figure}[!h]
\centering
    \includegraphics[width=\textwidth]{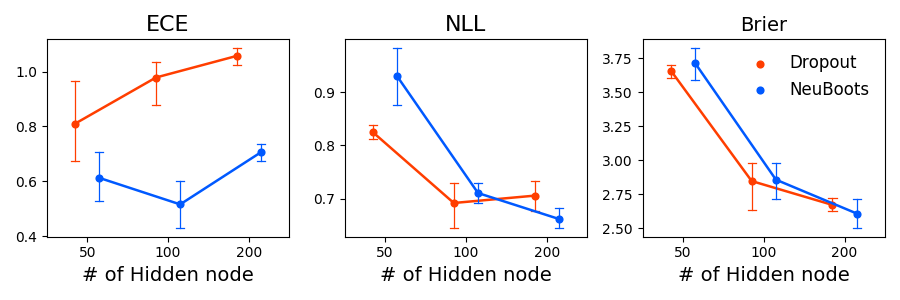}
    \caption{Comparison of ECE, NLL, and Brier for Dropout and the NeuBoots on the classification task on MNIST.}
    \label{fig:NeuBoots-Dropout-MNIST}
\end{figure}

\subsection{Calibration on Corrupted Dataset}
For evaluating the calibrated prediction of NeuBoots under  distributional shift, we use Corrupted CIFAR-10 and 100 datasets \citep{hendrycks2019benchmarking}.
Based on Ovadia et al.\citep{ovadia2019can}, we first train the ResNet-110 models on each training dataset of CIFAR-10 and 100, and evaluate it on the corrupted dataset.
For evaluation, we measure the mean accuracy and standard deviation for each of the five severities.

Deep Ensemble perform best in most cases, but NeuBoots also show better accuracy than baseline.

\begin{figure}[!h]
\centering
    \includegraphics[width=\textwidth]{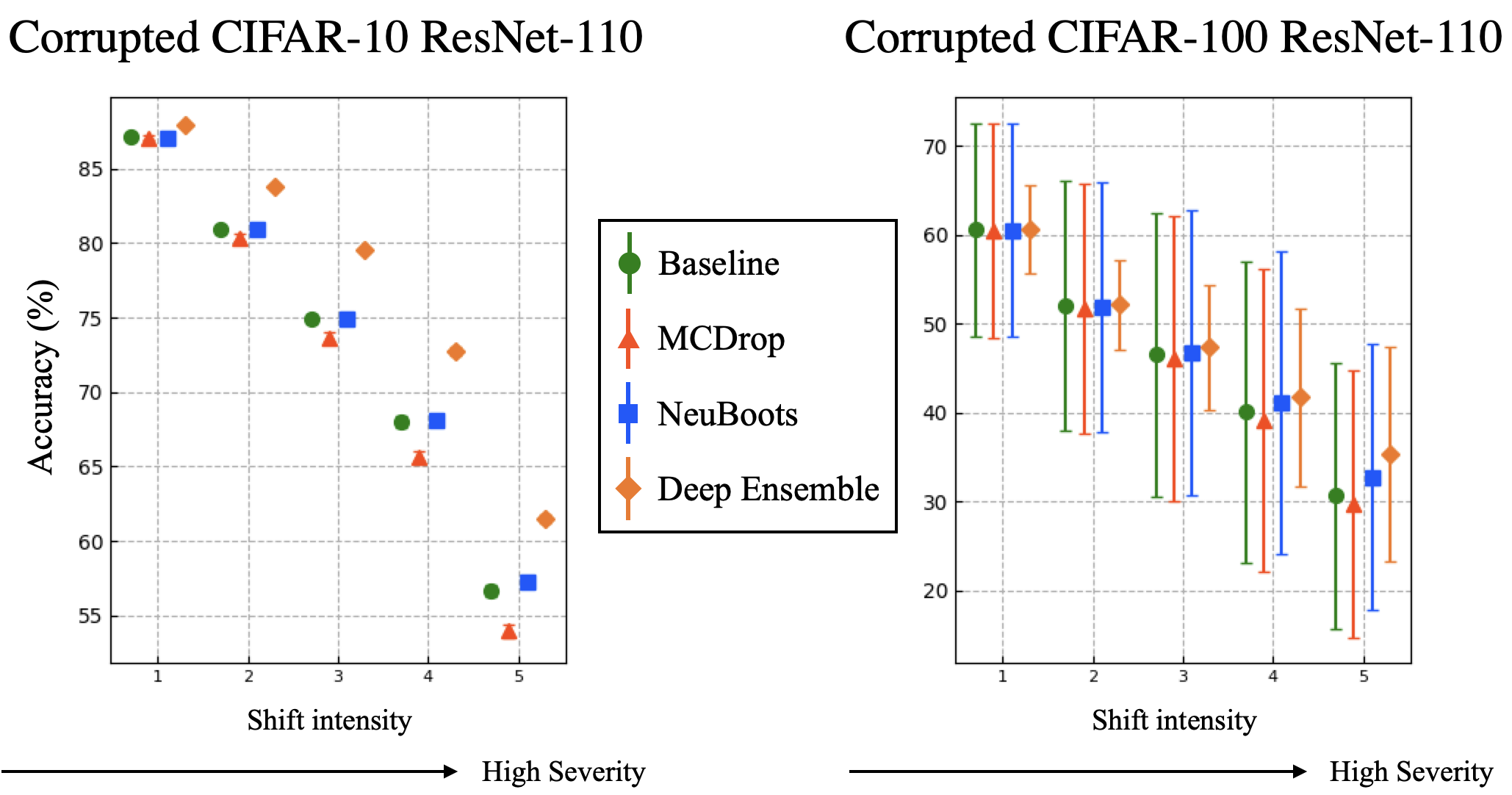}
    \caption{Calibration under distributional shift.}
    \label{fig:NeuBoots-Corrupted}
\end{figure}

\subsection{OOD Detection}

In this section, we illustrate additional results of OOD detection experiments. 

\begin{figure}[!h]
    \centering
    \includegraphics[width=0.7\textwidth]{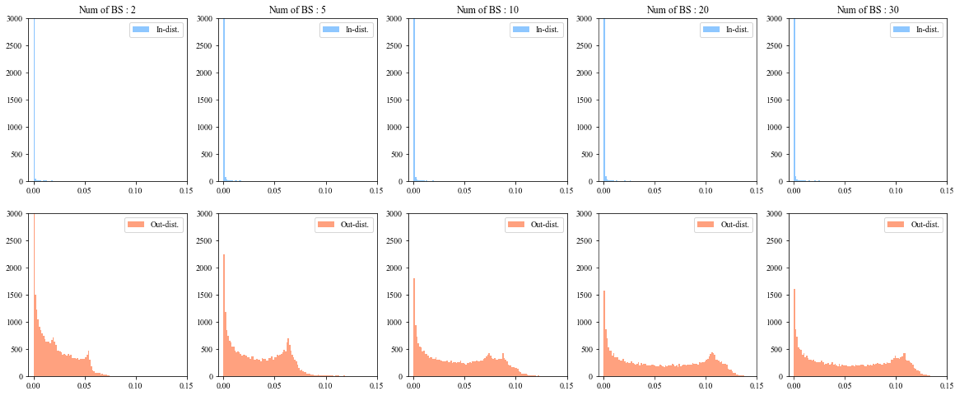}
    \caption{Histogram of the predictive standard deviation estimated by NeuBoots on test samples from CIFAR-10 (in-distribution) classes (top row) and SVHN (out-distribution) classes (bottom row), as we vary bootstrap sample size $B \in \{2, 5, 10, 20, 30\} $.  }\label{fig:ood-batch-size}
\end{figure}

\newpage

\begin{figure}[t]
    \centering
    \includegraphics[width=0.7\textwidth]{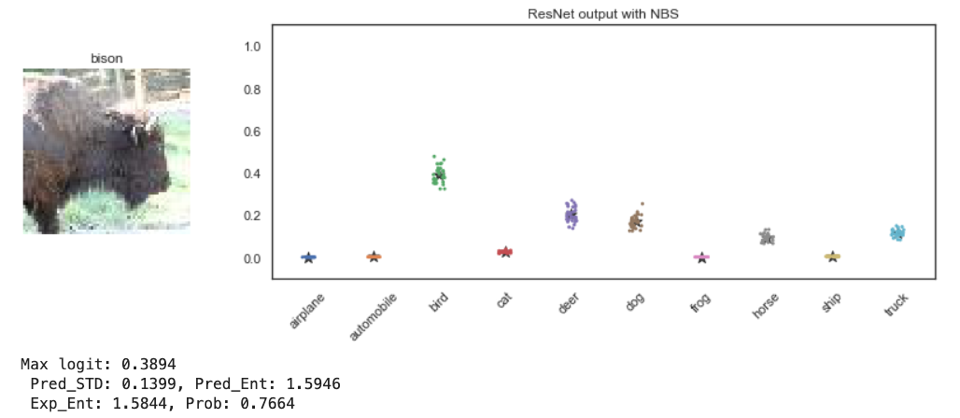}
    \caption{ Confidence bands of the prediction of NeuBoots for \cmtt{bison} data in TinyImageNet. The proposed method predicts is as an out-of-distribution class with \cmtt{prob}=0.7664.}
    \label{fig:qual1}
\end{figure}

\begin{figure}[t]
    \centering
    \includegraphics[width=0.7\textwidth]{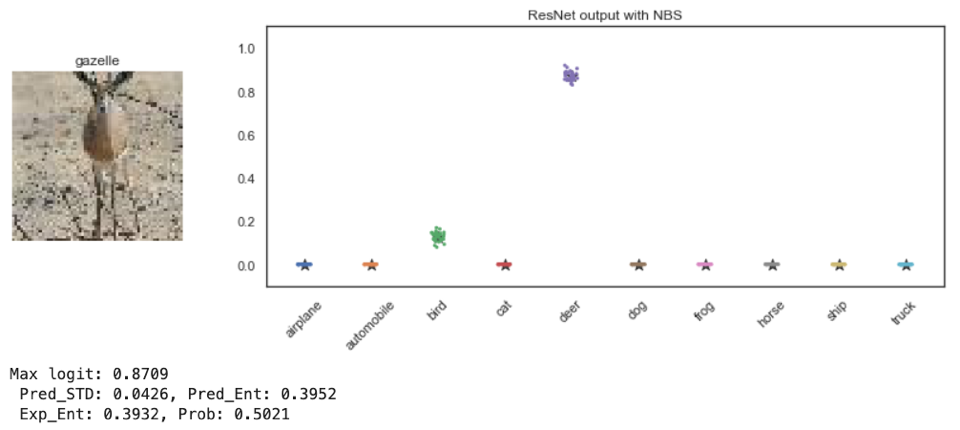}
    \caption{ Confidence bands of the prediction of NeuBoots for \cmtt{gazelle} data in TinyImageNet. The proposed method predicts is as an out-of-distribution class with \cmtt{prob}=0.5021. }
    \label{fig:qual2}
\end{figure}

\begin{figure}[t]
    \centering
    \includegraphics[width=0.7\textwidth]{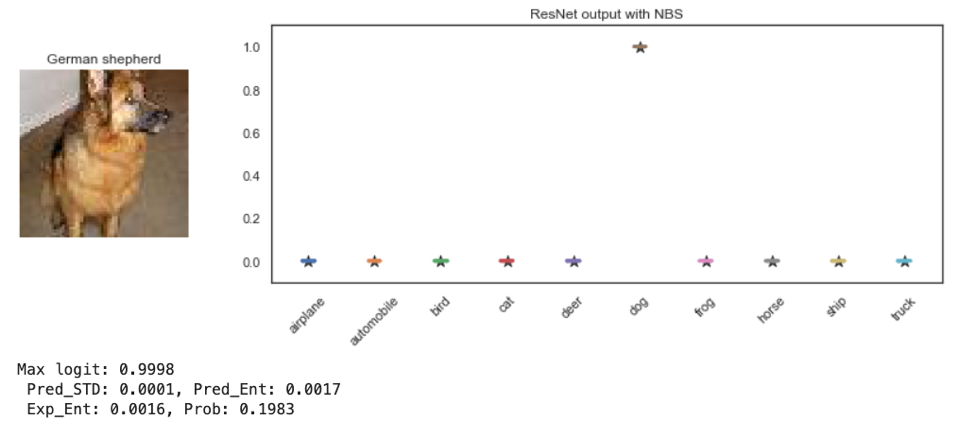}
    \caption{ Confidence bands of the prediction of NeuBoots for \cmtt{German shepherd} data in TinyImageNet. The proposed method predicts is as an in-of-distribution class \cmtt{dog} with \cmtt{prob}=0.1983. }
    \label{fig:qual3}
\end{figure}


\begin{thebibliography}{}

\bibitem[Aksela, 2003]{aksela2003comparison}
Aksela, M. (2003).
\newblock Comparison of classifier selection methods for improving committee
  performance.
\newblock In {\em International Workshop on Multiple Classifier Systems}, pages
  84--93. Springer.

\bibitem[Ashukha et~al., 2020]{Ashukha2020Pitfalls}
Ashukha, A., Lyzhov, A., Molchanov, D., and Vetrov, D. (2020).
\newblock Pitfalls of in-domain uncertainty estimation and ensembling in deep
  learning.
\newblock In {\em International Conference on Learning Representations}.

\bibitem[Breiman, 1996]{breiman1996bagging}
Breiman, L. (1996).
\newblock Bagging predictors.
\newblock {\em Machine learning}, 24(2):123--140.

\bibitem[Chen et~al., 2017]{chen2017rethinking}
Chen, L.-C., Papandreou, G., Schroff, F., and Adam, H. (2017).
\newblock Rethinking atrous convolution for semantic image segmentation.
\newblock {\em arXiv preprint arXiv:1706.05587}.

\bibitem[Choi et~al., 2021]{Choi2021.05.23.445351}
Choi, J., Kim, H.-J., Sim, G., Lee, S., Park, W.~S., Park, J.~H., Kang, H.-Y.,
  Lee, M., Heo, W.~D., Choo, J., Min, H., and Park, Y. (2021).
\newblock Label-free three-dimensional analyses of live cells with
  deep-learning-based segmentation exploiting refractive index distributions.
\newblock {\em bioRxiv}.

\bibitem[Cort{\'e}s-Ciriano and Bender, 2018]{cortes2018deep}
Cort{\'e}s-Ciriano, I. and Bender, A. (2018).
\newblock Deep confidence: a computationally efficient framework for
  calculating reliable prediction errors for deep neural networks.
\newblock {\em Journal of chemical information and modeling}, 59(3):1269--1281.

\bibitem[Efron, 1979]{efron1979bootstrap}
Efron, B. (1979).
\newblock Bootstrap methods: Another look at the jackknife.
\newblock {\em The Annals of Statistics}, 7(1):1--26.

\bibitem[Efron, 1987]{efron1987better}
Efron, B. (1987).
\newblock Better bootstrap confidence intervals.
\newblock {\em Journal of the American statistical Association},
  82(397):171--185.

\bibitem[Everingham et~al., 2010]{Everingham10}
Everingham, M., Van~Gool, L., Williams, C. K.~I., Winn, J., and Zisserman, A.
  (2010).
\newblock The pascal visual object classes (voc) challenge.
\newblock {\em International Journal of Computer Vision}, 88(2):303--338.

\bibitem[Fort et~al., 2020]{fort2020deep}
Fort, S., Hu, H., and Lakshminarayanan, B. (2020).
\newblock Deep ensembles: A loss landscape perspective.
\newblock {\em arXiv preprint arXiv:1912.02757}.

\bibitem[Franchi et~al., 2020]{franchi2020encoding}
Franchi, G., Bursuc, A., Aldea, E., Dubuisson, S., and Bloch, I. (2020).
\newblock Encoding the latent posterior of bayesian neural networks for
  uncertainty quantification.
\newblock In {\em NeurIPS workshop on Bayesian Deep Learning}.

\bibitem[Franke and Neumann, 2000]{franke2000bootstrapping}
Franke, J. and Neumann, M.~H. (2000).
\newblock Bootstrapping neural networks.
\newblock {\em Neural computation}, 12(8):1929--1949.

\bibitem[Gal and Ghahramani, 2016]{gal2016dropout}
Gal, Y. and Ghahramani, Z. (2016).
\newblock Dropout as a bayesian approximation: Representing model uncertainty
  in deep learning.
\newblock In {\em international conference on machine learning}, pages
  1050--1059.

\bibitem[Hall, 1986]{hall1986bootstrap}
Hall, P. (1986).
\newblock On the bootstrap and confidence intervals.
\newblock {\em The Annals of Statistics}, pages 1431--1452.

\bibitem[Hall, 1992]{hall1992bootstrap}
Hall, P. (1992).
\newblock On bootstrap confidence intervals in nonparametric regression.
\newblock {\em The Annals of Statistics}, pages 695--711.

\bibitem[Havasi et~al., 2021]{havasi2021training}
Havasi, M., Jenatton, R., Fort, S., Liu, J.~Z., Snoek, J., Lakshminarayanan,
  B., Dai, A.~M., and Tran, D. (2021).
\newblock Training independent subnetworks for robust prediction.
\newblock In {\em International Conference on Learning Representations}.

\bibitem[He et~al., 2016]{he2016deep}
He, K., Zhang, X., Ren, S., and Sun, J. (2016).
\newblock Deep residual learning for image recognition.
\newblock In {\em Proceedings of the IEEE conference on computer vision and
  pattern recognition}, pages 770--778.

\bibitem[Hendrycks and Dietterich, 2019]{hendrycks2019benchmarking}
Hendrycks, D. and Dietterich, T. (2019).
\newblock Benchmarking neural network robustness to common corruptions and
  perturbations.
\newblock {\em International Conference on Learning Representations}.

\bibitem[Hendrycks and Gimpel, 2017]{hendrycks17baseline}
Hendrycks, D. and Gimpel, K. (2017).
\newblock A baseline for detecting misclassified and out-of-distribution
  examples in neural networks.
\newblock {\em Proceedings of International Conference on Learning
  Representations}.

\bibitem[Huang et~al., 2017]{huang2017densely}
Huang, G., Liu, Z., Van Der~Maaten, L., and Weinberger, K.~Q. (2017).
\newblock Densely connected convolutional networks.
\newblock In {\em Proceedings of the IEEE conference on computer vision and
  pattern recognition}, pages 4700--4708.

\bibitem[Ioffe and Szegedy, 2015]{ioffe2015batch}
Ioffe, S. and Szegedy, C. (2015).
\newblock Batch normalization: Accelerating deep network training by reducing
  internal covariate shift.
\newblock In {\em International conference on machine learning}, pages
  448--456. PMLR.

\bibitem[Kim et~al., 2019]{kim2019scalable}
Kim, S., Kim, I., Lim, S., Baek, W., Kim, C., Cho, H., Yoon, B., and Kim, T.
  (2019).
\newblock Scalable neural architecture search for 3d medical image
  segmentation.
\newblock In {\em International Conference on Medical Image Computing and
  Computer-Assisted Intervention}, pages 220--228. Springer.

\bibitem[Kuleshov et~al., 2018]{kuleshov2018accurate}
Kuleshov, V., Fenner, N., and Ermon, S. (2018).
\newblock Accurate uncertainties for deep learning using calibrated regression.
\newblock In {\em International Conference on Machine Learning}, pages
  2796--2804.

\bibitem[Lakshminarayanan et~al., 2017]{lakshminarayanan2017simple}
Lakshminarayanan, B., Pritzel, A., and Blundell, C. (2017).
\newblock Simple and scalable predictive uncertainty estimation using deep
  ensembles.
\newblock In {\em Advances in neural information processing systems}, pages
  6402--6413.

\bibitem[Lee et~al., 2020]{lee2020bootstrapping}
Lee, J., Lee, Y., Kim, J., Yang, E., Hwang, S.~J., and Teh, Y.~W. (2020).
\newblock Bootstrapping neural processes.
\newblock {\em arXiv preprint arXiv:2008.02956}.

\bibitem[Lee et~al., 2018]{lee2018simple}
Lee, K., Lee, K., Lee, H., and Shin, J. (2018).
\newblock A simple unified framework for detecting out-of-distribution samples
  and adversarial attacks.
\newblock In {\em Advances in Neural Information Processing Systems}, pages
  7167--7177.

\bibitem[Liang et~al., 2018]{liang2018enhancing}
Liang, S., Li, Y., and Srikant, R. (2018).
\newblock Enhancing the reliability of out-of-distribution image detection in
  neural networks.
\newblock In {\em 6th International Conference on Learning Representations,
  ICLR 2018}.

\bibitem[Maddox et~al., 2020]{maddox2020simple}
Maddox, W.~J., Izmailov, P., Garipov, T., Vetrov, D.~P., and Wilson, A.~G.
  (2020).
\newblock A simple baseline for bayesian uncertainty in deep learning.
\newblock {\em Advances in Neural Information Processing Systems},
  32:13153--13164.

\bibitem[Moon et~al., 2020]{moon2020confidence}
Moon, J., Kim, J., Shin, Y., and Hwang, S. (2020).
\newblock Confidence-aware learning for deep neural networks.
\newblock In {\em international conference on machine learning}.

\bibitem[Naeini et~al., 2015]{naeini2015obtaining}
Naeini, M.~P., Cooper, G.~F., and Hauskrecht, M. (2015).
\newblock Obtaining well calibrated probabilities using bayesian binning.
\newblock In {\em Proceedings of the... AAAI Conference on Artificial
  Intelligence. AAAI Conference on Artificial Intelligence}, volume 2015, page
  2901. NIH Public Access.

\bibitem[Nalisnick and Smyth, 2017]{nalisnick2017amortized}
Nalisnick, E. and Smyth, P. (2017).
\newblock The amortized bootstrap.
\newblock In {\em ICML Workshop on Implicit Models.}

\bibitem[Newton and Raftery, 1994]{newton1994approximate}
Newton, M.~A. and Raftery, A.~E. (1994).
\newblock Approximate {B}ayesian inference with the weighted likelihood
  bootstrap.
\newblock {\em Journal of the Royal Statistical Society: Series B
  (Methodological)}, 56(1):3--26.

\bibitem[Ovadia et~al., 2019]{ovadia2019can}
Ovadia, Y., Fertig, E., Ren, J., Nado, Z., Sculley, D., Nowozin, S., Dillon,
  J.~V., Lakshminarayanan, B., and Snoek, J. (2019).
\newblock Can you trust your model's uncertainty? evaluating predictive
  uncertainty under dataset shift.
\newblock {\em Neural Information Processing System}.

\bibitem[Pr{\ae}stgaard and Wellner, 1993]{praestgaard1993exchangeably}
Pr{\ae}stgaard, J. and Wellner, J.~A. (1993).
\newblock Exchangeably weighted bootstraps of the general empirical process.
\newblock {\em The Annals of Probability}, pages 2053--2086.

\bibitem[Rame and Cord, 2021]{rame2021dice}
Rame, A. and Cord, M. (2021).
\newblock Dice: Diversity in deep ensembles via conditional redundancy
  adversarial estimation.
\newblock In {\em International Conference on Learning Representations}.

\bibitem[Reed et~al., 2014]{reed2014training}
Reed, S., Lee, H., Anguelov, D., Szegedy, C., Erhan, D., and Rabinovich, A.
  (2014).
\newblock Training deep neural networks on noisy labels with bootstrapping.
\newblock {\em arXiv preprint arXiv:1412.6596}.

\bibitem[Shao and Tu, 1996]{shao1996jackknife}
Shao, J. and Tu, D. (1996).
\newblock {\em The jackknife and bootstrap}.
\newblock Springer Science \& Business Media.

\bibitem[Shin et~al., 2020]{shin2020GBS}
Shin, M., Lee, Y., and Liu, J.~S. (2020).
\newblock Scalable uncertainty quantification via generative bootstrap sampler.
\newblock {\em arXiv preprint arXiv:2006.00767}.

\bibitem[{Smith} and {Gal}, 2018]{SmithGal2018Understanding}
{Smith}, L. and {Gal}, Y. (2018).
\newblock {Understanding Measures of Uncertainty for Adversarial Example
  Detection}.
\newblock In {\em UAI}.

\bibitem[Snelson and Ghahramani, 2006]{snelson2006sparse}
Snelson, E. and Ghahramani, Z. (2006).
\newblock Sparse gaussian processes using pseudo-inputs.
\newblock In {\em Advances in neural information processing systems}, pages
  1257--1264.

\bibitem[Vaswani et~al., 2017]{vaswani2017attention}
Vaswani, A., Shazeer, N., Parmar, N., Uszkoreit, J., Jones, L., Gomez, A.~N.,
  Kaiser, L., and Polosukhin, I. (2017).
\newblock Attention is all you need.
\newblock In {\em NIPS}.

\bibitem[Wen et~al., 2020]{wen2020batchensemble}
Wen, Y., Tran, D., and Ba, J. (2020).
\newblock Batchensemble: an alternative approach to efficient ensemble and
  lifelong learning.
\newblock In {\em International Conference on Learning Representations}.

\end{thebibliography}
\end{document}